\documentclass[3p,authoryear,preprint,10pt]{elsarticle}

\usepackage{lineno,hyperref}
\modulolinenumbers[5]

\journal{Journal of \LaTeX\ Templates}

\usepackage{comment}
\usepackage{float} 
\usepackage{subcaption}
\usepackage{array,booktabs}
\usepackage{adjustbox}
\usepackage{tabulary,xcolor}
\usepackage{amsfonts,amsmath,amssymb}
\usepackage{epsfig}
\usepackage[T1]{fontenc}
\makeatletter
\let\save@ps@pprintTitle\ps@pprintTitle
\def\ps@pprintTitle{\save@ps@pprintTitle\gdef\@oddfoot{\footnotesize\itshape \null\hfill\today}}
\def\hlinewd#1{%
  \noalign{\ifnum0=`}\fi\hrule \@height #1%
  \futurelet\reserved@a\@xhline}

\AtBeginDocument{\ifNAT@numbers \biboptions{sort&compress}\fi}
\makeatother

\usepackage{float}
\usepackage{ifluatex}
\ifluatex
\usepackage{fontspec}
\defaultfontfeatures{Ligatures=TeX}
\usepackage[]{unicode-math}
\unimathsetup{math-style=TeX}
\else 
\usepackage[utf8]{inputenc}
\fi 
\ifluatex\else\usepackage{stmaryrd}\fi

\usepackage{url,multirow,morefloats,floatflt,cancel,tfrupee}
\makeatletter

\AtBeginDocument{\@ifpackageloaded{textcomp}{}{\usepackage{textcomp}}}
\makeatother
\usepackage{colortbl}
\usepackage{xcolor}
\usepackage{pifont}
\usepackage[nointegrals]{wasysym}
\makeatletter

\def\mcWidth#1{\csname TY@F#1\endcsname+\tabcolsep}

\def\cAlignHack{\rightskip\@flushglue\leftskip\@flushglue\parindent\z@\parfillskip\z@skip}
\def\rAlignHack{\rightskip\z@skip\leftskip\@flushglue \parindent\z@\parfillskip\z@skip}

\@ifundefined{etal}{}{}

\usepackage{ifxetex}
\ifxetex\else\if@twocolumn\@ifpackageloaded{stfloats}{}{\usepackage{dblfloatfix}}\fi\fi

\AtBeginDocument{
\expandafter\ifx\csname eqalign\endcsname\relax
\def\eqalign#1{\null\vcenter{\def\\{\cr}\openup\jot\m@th
  \ialign{\strut$\displaystyle{##}$\hfil&$\displaystyle{{}##}$\hfil
      \crcr#1\crcr}}\,}
\fi
}

\AtBeginDocument{%
  \@ifpackageloaded{endfloat}%
   {\renewcommand\efloat@iwrite[1]{\immediate\expandafter\protected@write\csname efloat@post#1\endcsname{}}}{\newif\ifefloat@tables}%
}%

\def\BreakURLText#1{\@tfor\brk@tempa:=#1\do{\brk@tempa\hskip0pt}}
\let\lt=<
\let\gt=>
\def\processVert{\ifmmode|\else\textbar\fi}

\@ifundefined{subparagraph}{
\def\subparagraph{\@startsection{paragraph}{5}{2\parindent}{0ex plus 0.1ex minus 0.1ex}%
{0ex}{\normalfont\small\itshape}}%
}{}

\newcommand\role[1]{\unskip}
\newcommand\aucollab[1]{\unskip}
  
\@ifundefined{tsGraphicsScaleX}{\gdef\tsGraphicsScaleX{1}}{}
\@ifundefined{tsGraphicsScaleY}{\gdef\tsGraphicsScaleY{.9}}{}
\def\checkGraphicsWidth{\ifdim\Gin@nat@width>\linewidth
	\tsGraphicsScaleX\linewidth\else\Gin@nat@width\fi}

\def\checkGraphicsHeight{\ifdim\Gin@nat@height>.9\textheight
	\tsGraphicsScaleY\textheight\else\Gin@nat@height\fi}

\def\fixFloatSize#1{}
\let\ts@includegraphics\includegraphics

\def\inlinegraphic[#1]#2{{\edef\@tempa{#1}\edef\baseline@shift{\ifx\@tempa\@empty0\else#1\fi}\edef\tempZ{\the\numexpr(\numexpr(\baseline@shift*\f@size/100))}\protect\raisebox{\tempZ pt}{\ts@includegraphics{#2}}}}

\AtBeginDocument{\def\includegraphics{\@ifnextchar[{\ts@includegraphics}{\ts@includegraphics[width=\checkGraphicsWidth,height=\checkGraphicsHeight,keepaspectratio]}}}

\DeclareMathAlphabet{\mathpzc}{OT1}{pzc}{m}{it}

\def\URL#1#2{\@ifundefined{href}{#2}{\href{#1}{#2}}}

\def\UrlOrds{\do\*\do\-\do\~\do\'\do\"\do\-}%
\g@addto@macro{\UrlBreaks}{\UrlOrds}

\edef\fntEncoding{\f@encoding}

\makeatother

\newif\ifmultipleabstract\multipleabstractfalse%
\usepackage{natbib}

\begin{document}
\begin{frontmatter}

\title{Deep learning-based approaches for human motion decoding in smart walkers for rehabilitation}
    
\author[0001,0002]{Carolina Gon{\c{c}}alves} 
\author[0001,0002]{Jo{\~{a}}o M. Lopes}
\author[0003]{Sara Moccia}
\author[0004]{Daniele Berardini}
\author[0004]{Lucia Migliorelli}
\author[0001,0002,0005]{Cristina P. Santos}
\ead{cristina@dei.uminho.pt}

\address[0001]{Center for Microelectromechanical Systems (CMEMS)\unskip, 
    University of Minho\unskip, Guimarães, Portugal}
\address[0002]{LABBELS - Associate Laboratory\unskip, 
    Braga/Guimarães\unskip, Portugal}
\address[0003]{The BioRobotics Institute and Department of Excellence in Robotics and AI\unskip, Scuola Superiore Sant'Anna\unskip, Pisa, Italy}
\address[0004]{Department of Information Engineering\unskip,
Universit\`{a} Politecnica delle Marche\unskip, Ancona, Italy}
\address[0005]{Clinical Academic Center of Braga (2CA-Braga)\unskip, Braga, Portugal}


\begin{abstract}
Gait disabilities are among the most frequent impairments worldwide. Their treatment increasingly relies on rehabilitation therapies, in which smart walkers are being introduced to empower the user's recovery state and autonomy, while reducing the clinicians effort. For that, these should be able to decode human motion and needs, as early as possible. Current walkers decode motion intention using information gathered from wearable or embedded sensors, namely inertial units, force sensors, hall sensors, and lasers, whose main limitations imply an expensive solution or hinder the perception of human movement. Smart walkers commonly lack an advanced and seamless human-robot interaction, which intuitively and promptly understands human motions. A contactless approach is proposed in this work, addressing human motion decoding as an early action recognition/detection problematic, using RGB-D cameras. We studied different deep learning-based algorithms, organised in three different approaches, to process lower body RGB-D video sequences, recorded from an embedded camera of a smart walker, and classify them into 4 classes (stop, walk, turn right/left). A custom dataset involving 15 healthy participants walking with the walker device was acquired and prepared, resulting in 28800 balanced RGB-D frames, to train and evaluate the deep learning networks. The best results were attained by a convolutional neural network with a channel-wise attention mechanism, reaching accuracy values of 99.61\% and above 93\%, for offline early detection/recognition and trial simulations, respectively. Following the hypothesis that human lower body features encode prominent information, fostering a more robust prediction towards real-time applications, the algorithm focus was also quantitatively evaluated using Dice metric, leading to values slightly higher than 30\%. Promising results were attained for early action detection as a human motion decoding strategy, with enhancements in the focus of the proposed architectures.

\end{abstract}
    
  \end{frontmatter}
    
\section{Introduction}

In 2018, over a billion people were estimated to live with some form of disability \citep{WorldHealthOrganization}. This is caused by ageing population and by an increase in chronic health conditions \citep{WorldHealthOrganization}. Tumours, aneurysms, strokes and cerebellar ataxia are prominent causes of gait and posture impairments \citep{Mikolajczyk2018, Bonney2016, Jonsdottir2018, Celik2021}, which can have severe repercussions in strength, sensation, and movement, leading to lack of stability and increased risk of falls \citep{Mikolajczyk2018}. Rehabilitation therapies reveal promising results tackling these impairments \citep{Milne2017}. Gait rehabilitation requires long periods of intense physical exercise, presenting challenges for clinicians, due to the high demand of physical effort and inter-patient variability. Additionally, it may also be affected by the clinician experience and inter-clinician variability \citep{Mikolajczyk2018}, making this therapy more time-consuming and prone to error. To overcome these limitations, assistive technologies have emerged as effective means to increase the patient's independence and participation in their rehabilitation therapies \citep{WorldHealthOrganization2011}. Assistive technologies include Smart Walkers (SWs), which are intended to be used by or with humans. SWs no longer serve as just conventional physical supporters, but comprehend other intelligent functionalities to promote an efficient Human-Robot Interaction (HRI). These devices should be able to decode human motion and needs, as early as possible, which would be essential for a seamless HRI. This would be clinically relevant since it would enable a more natural and anticipated assistance, encouraging patients to take an active role in rehabilitation exercises or therapy sessions \citep{Zhao2020}. However, the interaction entailed by motion decoding should result from the device built-in sensors in order to maximise intuitiveness and technology acceptance.

Human motion decoding in SWs can be achieved through the use of wearable sensors, such as Inertial Measurement Units (IMU) \citep{Weon2018}. However, these have to be placed on the user's body, which hampers the clinician's task and the patient's movement, making rehabilitation more time consuming. Additionally, these can suffer electromagnetic interference from the walker's motors. Several sensors embedded in SWs have also been used for this purpose, although entailing other limitations that hinder rehabilitation. For instance, force sensors \citep{Cheng2017, Sierra2018} present a reduced long-term effectiveness \citep{Paulo2017}, infrared sensors \citep{Paulo2015} can be easily corrupted by light conditions and a handlebar specially designed with hall sensors \citep{Park2019} implies specific movements besides the natural gait, increasing the patient's cognitive load. Additionally, preprocessing may also be required to clean the output signals, for instance, when resorting to IMU or force sensors.

RGB-D cameras can be used along with SWs \citep{Palermo2021,9096121}. Although fostering a cheap, intuitive, and contactless solution, without interfering with the user's gait, these are not usually explored for the purpose of motion decoding. This can be due to the challenges inherent to image analysis, especially in realistic environments. Deep Learning (DL) methods have shown robustness tackling Human Action Recognition (HAR) \citep{Yeaser2020} and, more recently, Human Action Prediction (HAP) \citep{Chalen2019} from RGB-D images or videos. While the former tackles the issue of current action recognition or detection, the latter intends to anticipate the action's ending (early action recognition/detection), seeing only a small part of the action, or even its beginning (action anticipation), taking a step towards forecasting. Despite the progresses made, dealing with the hindrances of RGB-D image analysis, HAR and HAP methods have not yet been duly explored for human motion decoding in SWs.

Following the need of a more intuitive way of decoding motion in SWs, robust to realistic environments and capable of perceiving natural gait movements without hindering them, this work innovatively addresses motion decoding as a HAP problem. We propose the use of DL-based algorithms for online early detection and recognition of walking directions, from RGB-D videos. These videos, recorded by a SW embedded camera \citep{Lopes2021}, focus on the user's lower body, which we hypothesise to encode the most relevant motion information. Inspired by literature on HAP, five DL algorithms were implemented within three approaches. These approaches are responsible for implementing two different strategies: classification or segmentation-classification, in an attempt to evaluate and enhance the algorithm's focus on the human body, along with its performance. Different inputs from the RGB-D videos were evaluated, encoding the videos' temporal information into one single image. Therefore, this solution avoids not only extra sensors and a cognitive load to the user, but also computational-expensive complementary tasks, such as pose estimation, and additional computational complexity by not resorting to spatio-temporal DL models. To efficiently predict walking directions directly from camera streams, three key-performance requirements were also established to evaluate the DL algorithms:

\begin{itemize}
    \item \relax Accurately recognise and detect 4 actions (Stop, Walk, Turn Right (TR) and Turn Left (TL)), while only seeing small temporal sequences to increase the efficiency of the algorithm's response in online scenarios.
    \item \relax Extract human-centred features, when detecting these actions, to avoid background bias and motions as consequence of camera's movements. This will help to achieve a reliable performance for real-time applications, as the SW can only move after detecting the action.
    \item \relax Face online detection with a maximum admissible delay of 0.64s, which corresponds to the determined medium duration of one healthy step, while walking at the fastest velocity assumed by the walker (1m/s). This step time was determined in laboratory experiments and it is in accordance with \citet{Muller2017}.
\end{itemize}

This paper is organised as follows: Section \ref{sec:related_work} presents relevant related work, critically discussing its advantages and limitations. Section \ref{sec:methods} details all the methods for data acquisition and data preparation, along with the devised DL approaches, including model architectures and the involved algorithms. Section \ref{sec:experimental} summarises the implementation details for training and evaluation of the proposed models. Section \ref{sec:results} and Section \ref{sec:discussion} present and critically discuss all the obtained results, respectively. Finally, Section \ref{sec:conclusions} summarises the findings of this work and proposes future research insights.

\section{Related Work}
\label{sec:related_work}

\subsection{Human Motion Decoding in Smart Walkers}

Most of the current SWs decode human motion directly, demanding some level of physical intervention from the user and/or an extra load of cognitive effort. This direct mode has been typically implemented with specially designed handlebars, including force/pressure/load sensors \citep{Huang2005, Rodriguez-losada2008, Spenko2006, Jimenez2019, Cheng2017}, infrared cameras and Light Emitting Diodes \citep{Paulo2017} or Hall sensors \citep{Park2019}, which require specific hand motions to encode each walking direction. In an indirect mode, the walker becomes responsible for analysing the end-user's movement, inferring, from this, the walking directions. LIDAR sensors combined with wearable ones \citep{Weon2018} have been used to analyse the kinematics of lower limbs and measure feet orientation. \citet{Page2015} also resorted to a depth camera for feet position and orientation detection. \citet{LvL.YangJ.ZhaoD.2020} used multi-channel proximity sensors to determine each leg's distance and velocity. These studies highlighted the relevance of lower body features to naturally infer the user's walking directions. Nonetheless, the presented sensory technologies incur in additional costs for the motion detection task, including increased expenses (\emph{e.g.} LIDAR) and number of sensors, reducing technology acceptance (\emph{e.g.} IMU and proximity sensors), signal corruption by environmental conditions (\emph{e.g.} depth camera by sunlight) or electromagnetic interference, while causing long-term discomfort and hampering movement (\emph{e.g.} IMU).

Vision-based approaches should benefit the task of indirectly decoding human motion, avoiding the aforementioned limitations. Particularly, through the use of RGB-D cameras, usually built into SWs for other functionalities (\emph{e.g.,} gait and posture assessment \citep{Palermo2021}). \citet{Shen2020} used a RGB-D camera to perceive the intended walking direction from the torso kinematics. An hybrid Proportional-Integral-Derivative (PID) controller with integrated digital practical differentiator was then implemented to calculate the desired wheel rotation angles, being able to track the subject in forward and turning movements. However, the use of upper body information entailed a disadvantage: the upper body swaying \citep{Shen2020}. \citet{Chalvatzaki2019} complements RGB-D data with 2D laser data to perform real-time gait analysis, as well as to track the user's Centre-of-Mass position and velocity. This, assembled with the desired human-related robot coupling parameters, became the input used to forecast the human motion and the evolution of these parameters. The i-Walk system \citep{Chalvatzaki2020} points out the potential of using RGB-D cameras, in particular coupled with DL models, to attain an effective activity and gesture recognition, but not specifically to drive the walker. Although promising, these vision-based motion decoding approaches \citep{Shen2020, Chalvatzaki2019, Chalvatzaki2020} require additional computational tasks to estimate human poses or parameters from the acquired RGB videos. This can lead to error propagation, when feeding these inputs to a following motion decoding algorithm. 

To the best knowledge of the authors, no current SWs present a solution that decodes human motion directly from camera streams, decreasing the complexity inherent to human pose-based solutions, while enabling a seamless and intuitive HRI. This work contributes for this purpose by proposing DL-based approaches to early detect and recognise walking directions through lower-body RGB-D videos, while walking with a robotic walker.

\subsection{Vision-based DL Methods for HAR/HAP}

Recent studies have shown great potential for HAR \citep{Berardini2020, Li2016} and HAP \citep{Aliakbarian2017, Canuto2021, Gao2017}, resorting to DL algorithms fed with RGB-based inputs. Convolutional Neural Network (CNN) models, optionally combined with Long Short-Term Memory (LSTM), are the commonly used architectures in these tasks, usually resorting to video chunks or sequences of frames as input. The most common methods are based on colour (RGB) data \citep{Berardini2020}, on a combination of colour and depth (RGB-D) data \citep{Jalal2017}, or on skeleton data \citep{Li2016}.

Concerning HAR, \citet{Berardini2020} applied a DL solution to automatically recognise falls in stacks of seven RGB frames. The model architecture consisted on a CNN model as feature extractor, namely the VGG16, and a Bidirectional LSTM as feature classifier. \citet{Li2016} also implemented a LSTM model, aiming at Online Action Detection (OAD) through skeleton information. Moreover, simultaneously to frame-wise action (and background) classification, they proposed an estimation of the start and end frames of the current action, based on the definition of Gaussian-like curves for each action. As for \citet{DeGeest2018}, a different OAD proposal was made, using LSTM to model long term dependencies between actions, since human actions can imply a certain order (\emph{e.g.} standing, after being sited). However, these can also be random or environment-depending, as in free walking trajectories.
 
When handling HAP, there are two main lines of work in the literature: \textbf{i)} directly predicting the future frames' class or the current one, before the action ends (classification) \citep{Aliakbarian2017, Canuto2021, Guo2020, Ke2019} or \textbf{ii)} generating future visual representations that are further classified (regression followed by classification) \citep{Gao2017, Vondrick2016, Shi2018}. Moreover, a common focus of these studies is the design of novel loss functions that can reduce the predictive generalisation error \citep{Aliakbarian2017, Gao2017, Shi2018}.

\citet{Vondrick2016} resorted to visual representations as promising prediction targets. The proposed model consisted on AlexNet architecture, implementing five convolutional layers followed by five Fully Connected (FC) layers, with ReLu activations throughout the architecture. To deal with the model uncertainty, the last three FC layers presented \emph{K} networks, so each frame resulted in \emph{K} future visual representations, one for each plausible future. Approaches like \citet{Vondrick2016} anticipate only a representation of a fixed future time, based on a single past frame’s representation, dismissing the history and temporal trend. \citet{Gao2017} proposed instead a Reinforced Encoder-Decoder network which took multiple history representations as input and learned to anticipate a sequence of future ones. The video was segmented into chunks of six frames, each processed by a CNN model to extract a chunk representation. This was then fed to a LSTM-based Encoder-Decoder, outputting a prediction of the future video chunks' representations, in a supervised manner. The classification of these future representations was handled by two fully connected layers.

Despite all the research and developments, as well as the ability to use unlabelled videos for training, generating future representations is still defying, as well as time-consuming and prompt to error accumulation \citep{Ke2019}. Additionally, the learned representation may not be related to the action itself, as it can be influenced by background or other variables \citep{Aliakbarian2017}. Direct action prediction approaches avoid these limitations, by exploiting different forms of features and/or tailored losses to directly predict future classes.

\citet{Aliakbarian2017} resorted to the VGG16 model, connecting it to two different branches, one to compute context-aware features, encoding global information about the scene, and another to compute action-aware features. These features were then sequentially introduced in a multi-stage LSTM. The extraction of action-aware features relied on Class Activation Maps (CAMs) \citep{Zhou2016}, which indicate the regions in the input frame that most contributed to the class prediction. Nonetheless, this implied the training of CNN models before using them to compute CAMs, in an offline manner, which were then fed to the multi-frame classification LSTM model. Besides not being suitable for online applications, this use of CAMs does not guarantee these maps always focus on the action's relevant elements. For example, similar human activities may drive the model to extract features from the surroundings (\emph{e.g.} background movement, due to camera motion, objects, among others), as it is difficult to distinguish the human fine-grained movements. In \citet{Aliakbarian2017}, CAMs improvement was not taken into consideration during training, to assure the model's weights would be learnt in a way that would enhance action-centred feature extraction and improve the model's focus. These were just used, after training, for feature extraction.

In this sense, a new line of approaches is also emerging, namely the use of transformers \citep{Girdhar2019, Liu2019} and attention mechanisms \citep{Ke2019, Qiao2020, Wu2020}. Commonly for fine-grained action recognition, the frame or sequence of frames incorporates irrelevant or redundant information, with no discriminatory property. So, these algorithms guide the model to use \emph{attentional regions}, instead of the whole frame, enhance local features and attain selectively feature fusion. For example, \citet{Wu2020} implemented channel-wise and spatial attention mechanisms, along with baseline CNNs (VGG16 and ResNet-50) and LSTM. Additionally, when comparing to LSTM, transformers can be a lighter and maybe more suitable alternative for online performances \citep{Kozlov2020}. Nonetheless, the study of these algorithms for video analysis and HAR/HAP is still a newly tackled field.

Overall, the presented DL methods for HAR/HAP typically target a broad spectrum of actions (\emph{e.g.} blowing candles, falling, drinking), recorded with static cameras \citep{Berardini2020, Aliakbarian2017}, rather than specific walking directions (as stop, walking straight, turning right and left). The latter actions are the most used and relevant to specify and control the walker's trajectory \citep{Paulo2017, Park2019, Weon2018}, but demand no objects or interactions. The differences between them rely on human positions or fine-grained movements, incurring on a more challenging problem. Additionally, when resorting to mobile embedded cameras in SWs, the videos will include background motions and will be commonly recorded in a front-view perspective, hindering the perception of turning events.
   
\section{Methodology}
\label{sec:methods}

\begin{figure}[t]
    \centering
    \includegraphics[width=1\textwidth]{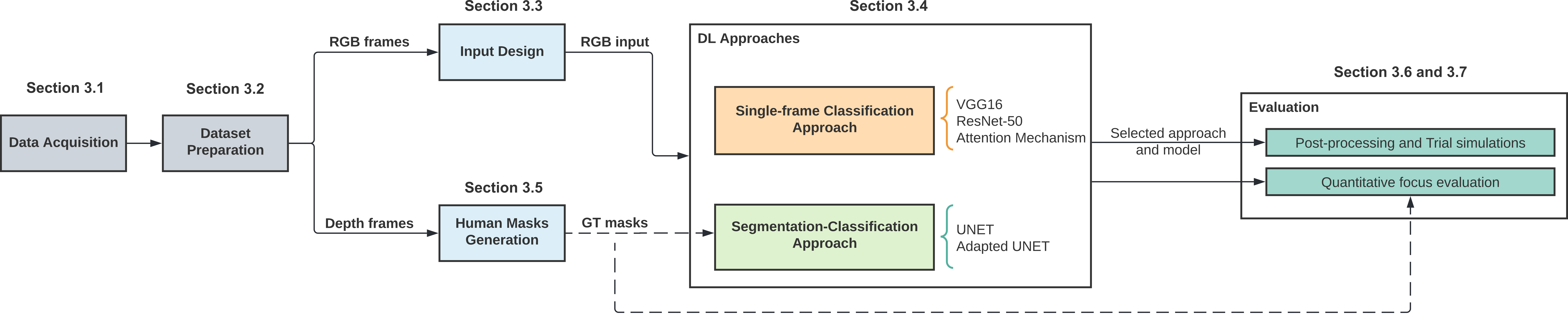}
    \caption{General workflow, including all the main methods implemented. GT stands for Ground-Truth.}
    \label{fig:methodology}
\end{figure}

This section describes the methodology followed in this work, from data acquisition and preparation, to the devised DL-based algorithms for model evaluation. Figure \ref{fig:methodology} illustrates the overall pipeline of the developed work and the inherent methodology.

\subsection{Data Acquisition}
\label{sec:acquisition}

Facing the lack of suitable public datasets for HAR or HAP with SWs, this work integrated the collection of a custom dataset, which included walking and turnings as different actions. This was recorded with mobile cameras of a SW, always maintaining a front-view perspective of the user. Data acquisition was conducted under the ethical procedures of the Ethics Committee in Life and Health Sciences (CEICVS 147/2021), following the Helsinki Declaration and the Oviedo Convention. All participants gave their informed consent to be part of the study. Data were collected at the School of Engineering of University of Minho, outside a controlled laboratory space.

\begin{figure}[t]
  \centering
  \includegraphics[width=0.5\textwidth]{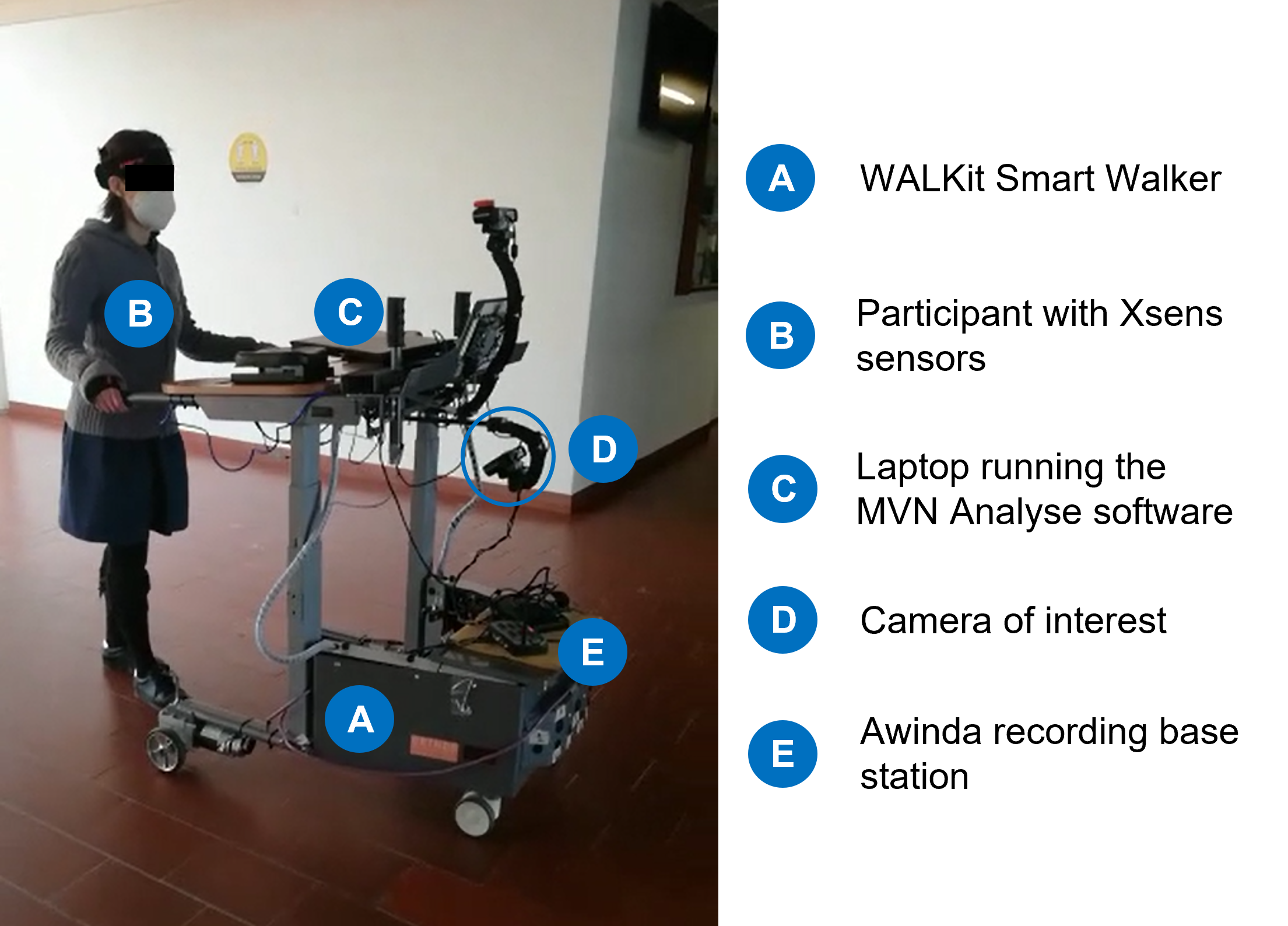}
  \caption{Mobile acquisition setup, where a laptop \textbf{(C)} running the Xsens MVN software and its acquisition base \textbf{(E)} are placed over the SW \textbf{(A)}, moving along with the robotic device. The user is equipped with the Xsens sensors \textbf{(B)}, hidden bellow a layer of clothing.}
  \label{fig:MobileAcquisitionSetup}
\end{figure}

A mobile acquisition setup was used for this purpose, as presented in \citet{Palermo2021}. This setup was composed by: \textbf{i)} WALKit Smart Walker (Figure \ref{fig:MobileAcquisitionSetup}A) used in rehabilitation \citep{Moreira2019,Lopes2021}, which integrates a RGB-D camera (Orbbec 3D Technology International Inc., USA) to assess the patient's gait, recording the user's legs and feet at a frame rate of 30 Hz; and \textbf{ii)} Xsens MVN Awinda (Xsens Technologies B.V., The Netherlands), which includes wearable IMUs (Figure \ref{fig:MobileAcquisitionSetup}B) and a base station (Figure \ref{fig:MobileAcquisitionSetup}E) connected to a laptop running the MVN software from Xsens (Figure \ref{fig:MobileAcquisitionSetup}C). This system recorded data at 60 Hz and has already been used before in other studies \citep{Palermo2021}. Its interest in this work was to provide relevant information to accurately place the Ground-Truth (GT) labels of the different actions, considering specific gait events (\emph{e.g.,} heel strike and toe-off \citep{Kurai2019}). Data synchronization was ensured by sending a digital pulse from WALKit SW to Xsens MVN  Awinda.

This mobile setup allowed for real-time data acquisition through pre-defined circuits, which are explained in Section \ref{sec:protocol_dataset}. Additionally, it enabled data to be recorded in realistic scenarios, with non-ideal light conditions and dynamic backgrounds to maximize data variability. 

\subsubsection{Dataset}
\label{sec:protocol_dataset}

The dataset consisted of RGB-D images acquired from fifteen healthy adults (nine males and six females), with average age of 25.27$\pm2.54$ years old, body mass of 65.15$\pm9.22$ kg and height of 168.47$\pm7.42$ cm.

Considering the aim of decoding human motion in SWs, the walking trajectories should be performed and recorded as naturally as possible, without causing abnormalities in the user's gait. WALKit SW allows to be used in a passive mode, but since it is heavy, this would produce non-natural slower and larger steps. Therefore, an automatic trajectory mode was developed, allowing the device to actively follow given trajectories, only as a recording device, creating velocity commands for each wheel according to the desired linear velocity for the SW and the turning strategy (angle and radius). This algorithm enabled the execution of four designed circuits, according to the turning direction (right or left) and curvature radius (wide and tight).

A circuit comprised a sequence of \textbf{i)} 10s standing, \textbf{ii)} 3-meter walking, \textbf{iii)} 90º turn, \textbf{iv)} 3-meter walking, and \textbf{v)} 10s standing. Each participant performed 2 valid trials per circuit and per gait speed: 0.5 m/s, 0.7 m/s, and 1 m/s. These speeds were selected according to the walker's most commonly used velocity range, as well as considering the typical self-selected slow, normal, and fast walking speeds for healthy subjects \citep{OCallaghan2020}. Different light conditions were held for each trial, increasing the environment and movement variability (\emph{e.g.} backgrounds, lighting, velocities, turn features), while improving the statistical significance of the data. 

The circuit sequence remained unknown to the participants until the beginning of the trial, when a brief explanation was given. The floor was marked with tape and signalised, during the recordings, with chairs and staff people, so that participants could see and react to the circuit's morphology. The marked spots were positioned in a way that remained invisible to the SW's cameras, during the performance of each respective trial. Additionally, the overall circuit and trial’s order was randomised for each subject before the beginning of the acquisition.

The participants were instructed not to just follow the smart walker, but to interact with it, along the designed and marked circuit. This enabled realistic scenarios, while capturing the user's intention as naturally as possible. Some familiarisation trials were also performed before the real recording procedure, encouraging the HRI. Overall, each subject performed 24 trials, taking no more than 1 minute each.

The labelling process was executed in real-time, along the data acquisition, in two different ways: \textbf{i)} with joystick commands and \textbf{ii)} with velocity commands. The former relied on an external person using joystick digital buttons, similar to \citet{Figueiredo2020}, to mark transitional moments between actions, according to the observation of clear feet movements. This produced GT labels responsible for denoting the user's interaction and intention (\emph{JOY labels}). The latter represents the device's actions, generating GT labels mainly for action recognition, when the background is already changing accordingly to the performed movements (\emph{VEL labels}).

\subsection{Dataset Preparation}
\label{sec:dataset_prep}

Since the principal walking frequency is no higher than 2 Hz for gait speeds above 1 m/s \citep{Pachi2005FREQUENCYAV}, all data was down-sampled to 15 Hz, covering more time of action with a lower number of similar samples. Aiming the creation of a balanced dataset, 40 frames were extracted from each video sequence. This number was selected considering the minimum number of frames obtained for the turn actions during the trials performed by the participants. This extraction was performed avoiding the action's boundaries or transitions to prevent possible bias in the labelling procedure, while aiming an effective early action recognition. For this reason, the start of each class was marked with the latest of the two labels (\emph{JOY} and \emph{VEL}). In total, the final dataset contained 28800 RGB-D frames. 

To simulate real scenarios (Section \ref{sec:post_processing}), a dataset containing the data of the test participants was created, including the complete RGB trial videos, towards online early action detection. The foot contacts obtained with Xsens MVN  Awinda were used to better position the \emph{JOY} labels, enabling more accurate labels to mark action transitions.

\subsection{Input Design}

Literature revealed the use of windows of frames \citep{Berardini2020} and/or other computed inputs, as the optical flow \citep{Simonyan2014}. This paper proposes two types of input, computed within a sliding window over temporal video clips: \textbf{i)} the difference between the last and first RGB frames (\emph{DIF} input) and \textbf{ii)} the sum of all RGB images (\emph{ADD} input), in each temporal window.

\begin{figure}[t]
    \centering
    \includegraphics[width=0.7\textwidth]{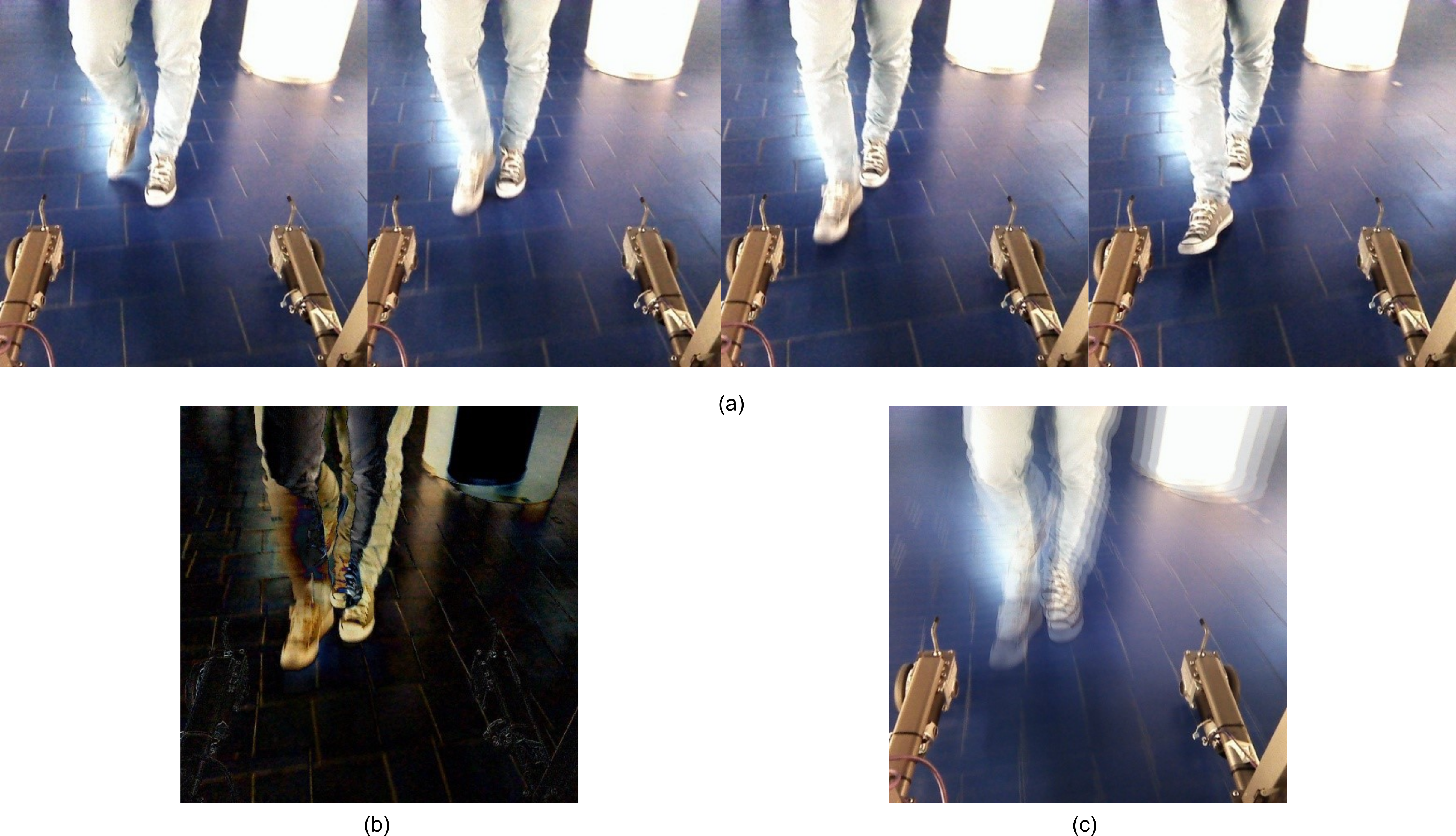}
    \caption{\textbf{(a)} A TR sequence extracted from our dataset. \textbf{(b)} \emph{DIF} image, corresponding to the difference between the last and the first window frames. \textbf{(c)} \emph{ADD} image, computed from the sum of all window frames.}
    \label{fig:dif_add_vs_originals}
\end{figure}

The window length was empirically defined as 4 (0.27s), since it incorporates visible human motion across all gait speeds, while never containing information from more than 1 different step. Examples to better understand the correlation between the original frames and the resultant \emph{DIF} and \emph{ADD} images are presented in Figure \ref{fig:dif_add_vs_originals}.

The designed inputs were optionally cropped, removing surrounding background, in an attempt for the model to direct its focus to the user and his/hers fine-grained movements. The cropping procedure was performed according to a Region of Interest (ROI), instead of using bounding box localization networks, since the user constantly assumes a central position in the image (in the middle of the walker's handles). In total, four input forms were generated and studied in this work to recognise and detect human motions: cropped and non-cropped \emph{DIF} and \emph{ADD} images.

\subsection{Deep-Learning Approaches}
\label{sec:DL}

\begin{figure}[t]
    \centering
    \includegraphics[width=0.6\textwidth]{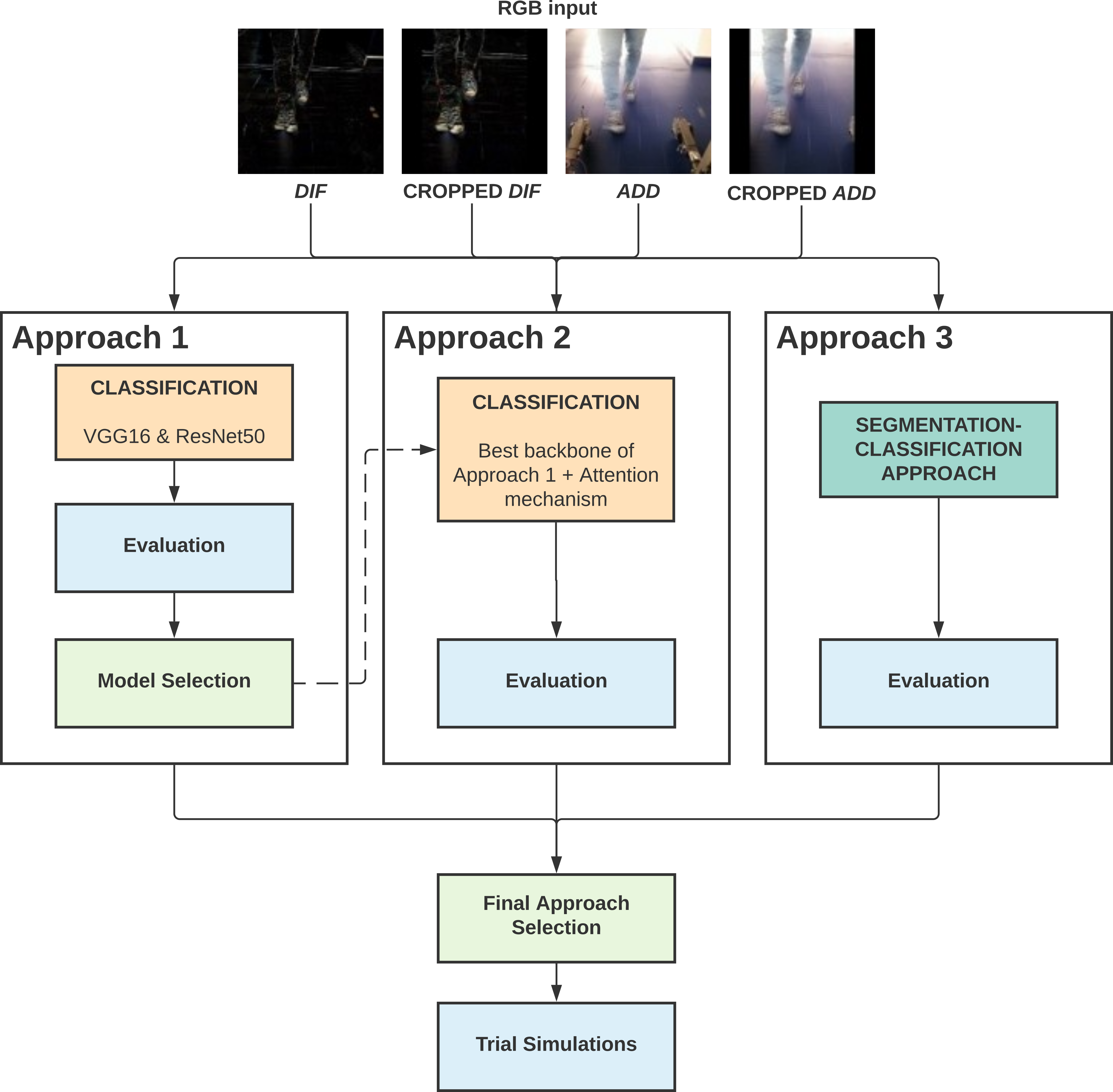}
    \caption{Workflow of the devised approaches, showing how they were conducted.}
    \label{fig:workflow}
\end{figure}

In this work, we have implemented three different approaches (\textbf{Approach 1}, \textbf{Approach 2}, and \textbf{Approach 3}), as illustrated in Figure \ref{fig:workflow}, which were trained and evaluated considering the four proposed input forms (\textit{i.e.}, cropped and non-cropped \emph{DIF} and cropped and non-cropped \emph{ADD} images).

\begin{figure}[t!]
    \centering
    \includegraphics[width=1.0\textwidth]{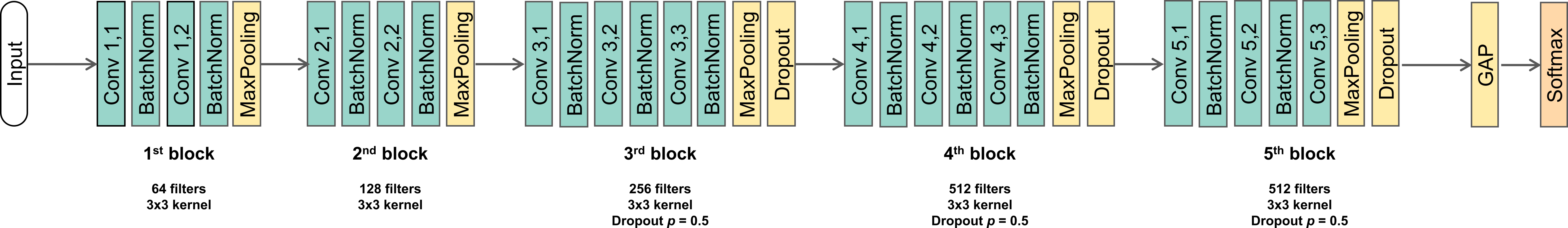}
    \caption{Diagram of the implemented VGG16 architecture.}
    \label{fig:vgg16_design}
\end{figure}

Regarding \textbf{Approach 1}, this consisted on single-frame classification. Here, the VGG16 model was used without its top layers \citep{Berardini2020}, only with a global average pooling (GAP) followed by a softmax classification layer, as it is illustrated in Figure \ref{fig:vgg16_design}. Batch normalization layers were also added to improve the optimization and generalization performance. Moreover, the activation function of the last convolution layer was changed from ReLU to hyperbolic tangent function (\emph{tanh}), to decrease the clipping of final feature map values and avoid possible vanishing gradients. The ResNet-50 model was also used, considering its configuration presented in \citet{He2016}.

\begin{figure}[t]
    \centering
    \includegraphics[width=0.5\textwidth]{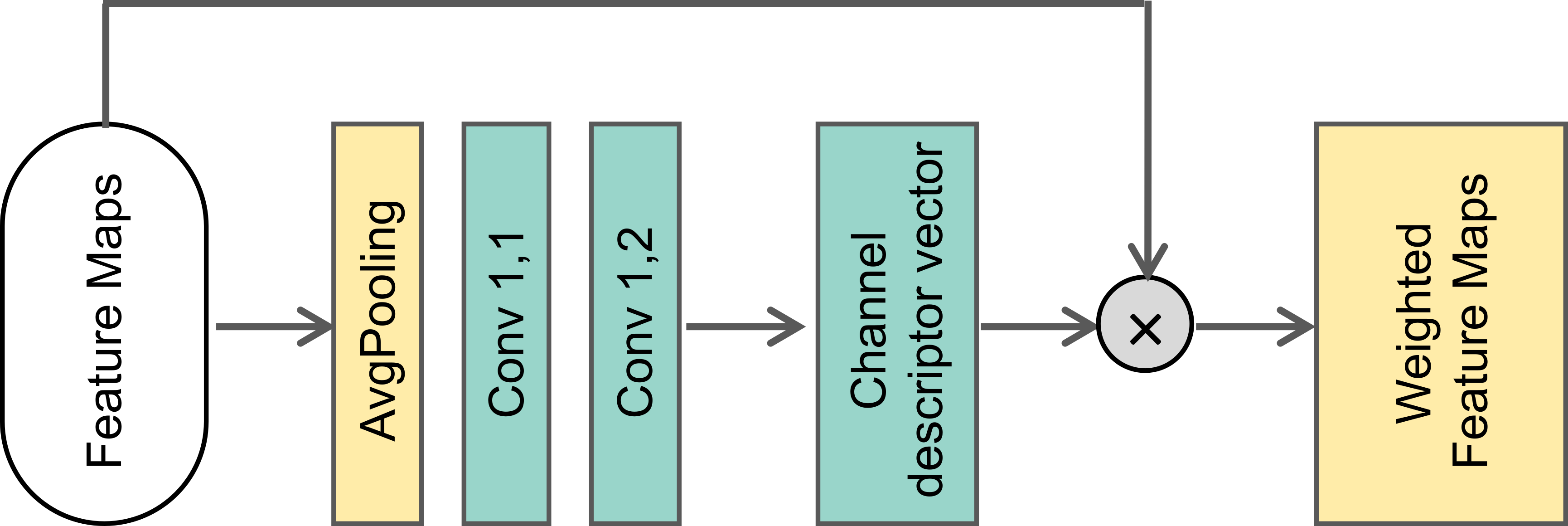}
    \caption{Diagram of the implemented channel-wise attention mechanism as presented in \citet{Wu2020}. The channel-attention weighted feature maps are then fed to the Global Average Pooling of the selected CNN model.}
    \label{fig:att_mech_design}
\end{figure}

\textbf{Approach 2} consisted of the best backbone obtained in \textbf{Approach 1} with a channel-wise attention mechanism, as presented in \citet{Wu2020}. This was tested as an attempt to increase the model focus towards the user’s legs and feet. The attention mechanism follows the network's last convolutional block and creates a channel descriptor vector with the corresponding weights for each feature map. This vector was computed through an average pooling and two convolutional layers, the first followed by a ReLU activation function and the second by a sigmoid function. The obtained vector is then used to perform a multiplication between each feature map and the corresponding computed weight, as it is illustrated in Figure \ref{fig:att_mech_design}.

\begin{figure}[t]
    \centering
    \includegraphics[width=0.5\textwidth]{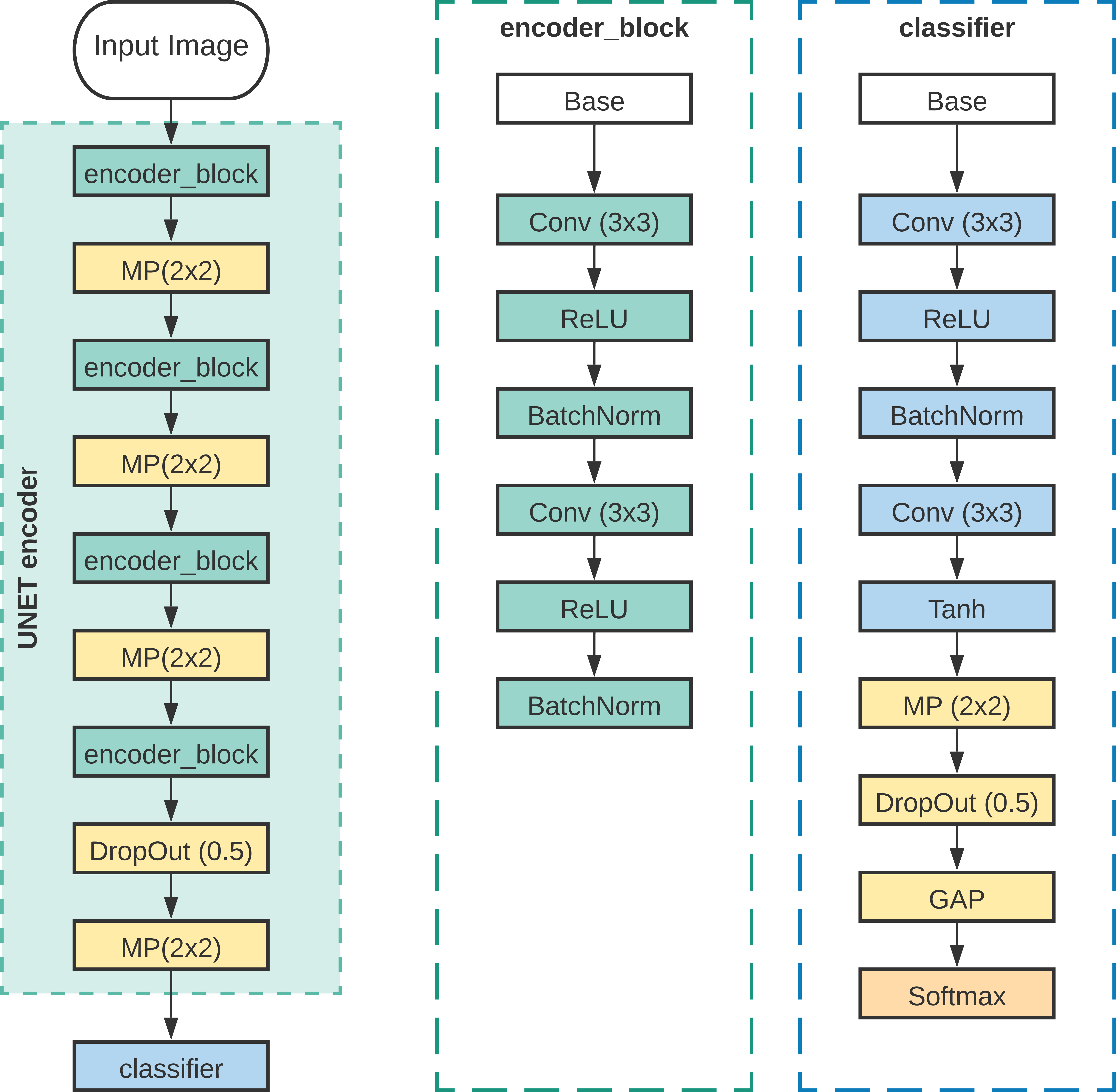}
    \caption{Schematic of the adapted UNET model for single-frame classification. MP stands for MaxPooling, GAP stands for Global Average Pooling, and the tuples of numbers represent the respective kernel sizes and 0.5 the dropout rate.}
    \label{fig:unet_for_classification}
\end{figure}

As for \textbf{Approach 3}, UNET neural network \citep{Ronneberger2015} was used to segment the user’s legs and feet in order to obtain the optimized weights which could allow the neural network to focus on that region of the user’s body. These optimized weights were used to pre-train a novel classification model. This latter model integrates the UNET encoder, followed by two convolutional blocks. A dropout layer with probability of 50\% was added, followed by a GAP and a softmax layers, decoding human motion into the four classes (stop, walk, TR, and TL). Figure \ref{fig:unet_for_classification} illustrates the neural network used as the basis of \textbf{Approach 3}. 

As presented in \citet{Berardini2020}, transfer learning technique was used for all approaches. The backbone weights of VGG16 and ResNet-50 classification models were initialised to the weights of these same state-of-the-art networks pre-trained on the general object classification benchmark, ImageNet \citep{Deng2010}, towards accuracy maximisation, while using large amounts of computational resources, which trains the models to detect and produce general features. In the segmentation-classification approach, the UNET encoder weights of the classification model were initialised with the ones learned in the previous training of the UNET model for segmentation, while the remaining layers were initialised with the He normal function. The first 16 layers of this encoder were also frozen during classification training.

\subsection{Human Masks Generation}
To train the UNET neural network, human masks were computed through an adaptation of our algorithm presented in \citet{9096121}, using depth frames. This algorithm involved geometric and threshold operations that removed the background, as well as the floor plane, to isolate the user. Depth frames can be corrupted by the high exposure from sunlight, which, in extreme conditions, causes unrecognisable silhouettes in the computed masks. These corrupted masks were duly removed through empirically established thresholds. When facing minor corruptions (\emph{e.g.,} incomplete feet), the masks were corrected through classic vision algorithms. These were processed accordingly to form the final GT masks, corresponding to the input used for model training and evaluation (cropped or non-cropped \emph{DIF} or \emph{ADD}). Examples of the obtained masks can be found in \ref{appendix_seg}.

\subsection{Quantitative focus evaluation}
Due to camera motion, background movements are also present in the input images (Figure \ref{fig:dif_add_vs_originals}), according to the action performed by the SW. This can compromise early action detection in future real-time applications, as the model should be focused on lower body features to predict walking directions, specially during their first occurrence, without overfitting to the background.

An algorithm for consistent quantitative evaluation of this focus was implemented. Following \citet{Selvaraju2019}, grad-CAMs heatmaps were generated for each input image, over the model last convolutional layer. These were then compared with the respective GT human masks, also used as segmentation labels. Hence, the model focus was evaluated through the similarity between the grad-CAMs heatmaps and the GT masks.

\subsection{Trial simulations and Post-processing}
\label{sec:post_processing}

The results from each approach were analysed and compared, towards the selection of the most promising model architecture, and input form. These were then evaluated in trial simulations, assessing the performance in online early action detection tasks.

To deal with the model uncertainty while predicting, a post-processing technique was designed and implemented. This technique assumes a minimum action duration (2s, which equals to, at least, two complete steps in every considered gait speed) and applies it to the previous predicted class, meaning that it only allows for an action transition to happen if the previous predicted one lasted, at least, 2s. Additionally, it also reduces the spectrum of possible transitions by not allowing a TR/TL to happen immediately after a TL/TR event, respectively. These conditions help avoiding prediction errors and are consistent with rehabilitation therapy sessions. This post-processing technique also reduces possible on-off noise, without introducing delays in the decision process.

\section{Experimental Protocol}
\label{sec:experimental}

This section details the experimental protocol to train and evaluate the proposed approaches, including the dataset split and the training hardware used.

\subsection{Dataset Split}
\label{sec:dataset_details}

The dataset split followed the 80-20 split configuration: 80\% was used for training (12 participants) and the remaining 20\% was used for testing (3 participants). To control the training process, one participant was left to the validation set, similar to \citet{Canuto2021}. Table \ref{tab:dataset_split} describes each of these splits. For the tasks involving the generated human masks, namely segmentation and the quantitative focus evaluation, the dataset used was smaller, since some sequences were removed giving mask corruption reasons.

To perform the trial simulations, the test split was used to create a new dataset, as mentioned in Section \ref{sec:dataset_prep}. In total, this test dataset included 72 untrimmed videos from 3 participants (8, 11, 15, according to Table \ref{tab:dataset_split}).

\begin{table}[H]
\centering
\caption{Constitution of each dataset split}
\label{tab:dataset_split}
\maxsizebox{0.6\textwidth}{!}{%
\begin{tabular}{@{}cccc@{}}
\toprule
\multirow{2}{*}{\textbf{Split}} & \multirow{2}{*}{\textbf{Participants IDs}} & \multicolumn{2}{c}{\textbf{Number of images}} \\ \cmidrule(l){3-4} 
 &  & \textbf{Classification} & \textbf{Segmentation} \\ \midrule
Train & [1,15) $\setminus \{5, 8, 11\}$ & 19536 & 13209 \\
Validation & 5 & 1776 & 1295 \\
Test & 8, 11, 15 & 5328 & 4736 \\ \bottomrule
\end{tabular}%
}
\end{table}

\subsection{Training Details}
\label{sec:implementation_details}

\textbf{CNN configuration:} Reasonable hyperparameters were extracted from literature that tackles similar tasks and models (\citet{Aliakbarian2017} for single-frame classification and \citet{Ronneberger2015} for segmentation). For classification, the parameters were optimised using Mini Batch Gradient Descent, with learning rate of 0.001, Nesterov momentum of 0.9 and batch size of 64. The initial learning rate was decayed by 50\% until a minimum of 1$e^{-4}$, if the training loss did not improve within 4 epochs. As loss function, the categorical cross-entropy was used. For segmentation, the Adam optimizer was used instead, with a learning rate of 1$e^{-4}$ and a batch size of 16. UNET convolution layers were initialised with He normal weights initialization. As loss function, the binary cross-entropy was used. At the end of the training, the best model was selected according to the validation F1-score or loss, for classification and segmentation, respectively.

\textbf{Data preparation:} The generated input images were resized, from a resolution of 480x480 pixels to a resolution of 224x224 pixels, preserving the images' aspect ratio, to match the pre-trained networks' input size. For the cropped inputs, the original aspect ratio decreased with the cropping procedure and the image was thus resized as much as possible, towards the target resolution. After this, padding was performed, centring the image, to fulfil the 224x224 dimensions. Image augmentation was then used in order to avoid overfitting, as well as to improve the model focus on the human body, despite its global position in the images. Random alterations were applied to the image brightness and contrast, as well as spatial augmentations, such as height and width shifts and zoom. Since directions are an important feature to distinguish between TR, TL and straight walking, rotations were avoided. Additionally, random Gaussian blur was added. Finally, the images where normalised between 0 and 1. 

\textbf{Training Hardware:} All models were developed offline, with the collected data, using the Tensorflow and Keras DL library, on a Python environment. The depicted experiments and DL approaches were trained in a computer with the following hardware specifications: \textbf{GPU}: 1x Nvidia GeForce RTX 3080 Ti/PCle/SSE2; \textbf{CPU}: Intel(R) Core(TM) i9-10940X @3.3GHz (14 core, 28 threads); \textbf{RAM}: ~65,5 GB.

\subsection{Performance Metrics}
\label{sec:metrics}

Classification was evaluated using common metrics for this task, namely accuracy (ACC), precision, recall, and F1-score \citep{Berardini2020, Aliakbarian2017}. These metrics are defined in the equations below, where \emph{TP}, \emph{TN}, \emph{FP}, \emph{FN} refer to the true/false positive/negative observations and \emph{n} stands for the total number of classes (in this case, 4).

\begin{equation} 
    \mathrm{ACC} = \sum_{i=1}^{n}\frac{TP_i+TN_i}{TP_i+TN_i+FP_i+FN_i}
\end{equation}
\begin{equation}
    \mathrm{Precision} = \frac{1}{n}\sum_{i=1}^{n}\frac{TP_i}{TP_i+FP_i}
\end{equation}

\begin{equation}
    \mathrm{Recall} = \frac{1}{n}\sum_{i=1}^{n}\frac{TP_i}{TP_i+FN_i}
\end{equation}

\begin{equation}
    \mathrm{F1-score} = \frac{2*\mathrm{Precision} \times \mathrm{Recall}}{\mathrm{Precision}+\mathrm{Recall}}
    \end{equation}

Intersection over Union (IoU) and Dice (Dice) were computed for segmentation and quantitative focus evaluation, as shown by the following equations (\emph{TP}, \emph{FP} and \emph{FN} refer to the true positive and false positive/negative observations).

\begin{equation} 
   \mathrm{IoU} = \frac{TP}{TP + FP + FN}
\end{equation}

\begin{equation}
    \mathrm{Dice} = \frac{2 \times TP}{(TP + FP) + (TP + FN)}
\end{equation}

Inspired by \citet{Baptista-Rios2020}, OAD metrics were implemented to better evaluate the performance in online simulations, such as instantaneous accuracy (IA), instantaneous precision (IP), instantaneous weighted accuracy (wIA) and instantaneous calibrated precision (cIP). These are represented in the following equations, were \emph{t} corresponds to the time instant, \emph{K} to the the total population considered until time \emph{t}, \emph{n} to the number of classes and \emph{TP}, \emph{TN}, \emph{FP}, \emph{FN} refer to the seen true/false positive/negative observations overall classes.

\begin{minipage}[h!]{0.4\textwidth}
\begin{flalign}
    \mathrm{IA} &= \frac{1}{K \times n} \sum_{j=0}^{t} TP_j + TN_j &&
\end{flalign}
\vspace{-0.5cm}
\begin{flalign} 
    \mathrm{IP} &= \frac{\sum_{j=0}^{t} TP_j}{\sum_{j=0}^{t} TP_j+FP_j}&&
\end{flalign}
\end{minipage}
\hspace{0.5cm}
\begin{minipage}[H]{0.5\textwidth}
\begin{flalign} 
    \mathrm{wIA} &=  \frac{1}{K \times n} [ w \times \sum_{j=0}^{t} TP_j + \frac{1}{w} \times \sum_{j=0}^{t} TN_j ] &&
\end{flalign}
\vspace{-0.5cm}
\begin{flalign} 
    \mathrm{cIP} &= \frac{w \times \sum_{j=0}^{t} TP_j}{w \times \sum_{j=0}^{t} TP_j+ \sum_{j=0}^{t} FP_j}&&
\end{flalign}
\end{minipage}
\vspace{0.5cm}

These metrics evaluate the model performance as the frames are acquired, without having to wait to an unknown end. Additionally, wIA and cIP condition the value of a true observation to the total negatives \emph{vs.} total positives ratio (\emph{w}), which is dynamic and always based only on the seen portion of the video.

\section{Results}
\label{sec:results}

Starting from the CNN models for single-frame classification (\textbf{approach 1}), to an attention mechanism (\textbf{approach 2}) and the segmentation-classification approach (\textbf{approach 3}), the influence of the various forms of input is studied and the classification models are evaluated in two aspects: \textbf{i)} the accuracy of the predicted labels; and \textbf{ii)} model focus on the human body region.

\subsection{Single-frame Classification Approaches }
\label{sec:results_segmentclassif}

Table \ref{tab:input_val_results} shows the validation results, obtained for \textbf{approaches 1} and \textbf{2} considering single-frame classification. The respective training curves can be found in \ref{appendix_classif}. According to Table \ref{tab:input_val_results}, ResNet-50 outperformed the VGG16, except for the cropped \emph{DIF} input. The \emph{DIF} revealed here a better performance, followed by the \emph{ADD} input, which was only worst by an overall maximum margin of approximately 2\%.

Adding a channel-wise attention mechanism to ResNet-50 further enhanced its classification performances, specially for the cropped inputs. The F1-score was increased by 3.98\% and 3.85\%, for the cropped \emph{DIF} and \emph{ADD} inputs, respectively. This mechanism also changed the influence exerted by the different types of images. With this model, \emph{ADD} input surpassed the accuracy and F1-score values obtained for the \emph{DIF} input, by a maximum margin of 1.3\%.

\begin{table}[H]
\centering
\caption{Validation results of the VGG16 and ResNet-50 (without and with a channel-wise attention mechanism), as well as the training time}
\label{tab:input_val_results}
\maxsizebox{0.8\textwidth}{!}{%
\begin{tabular}{ccccccc}
\hline
\textbf{Input Type} & \textbf{Crop} & \textbf{ACC (\%)} & \textbf{F1-score (\%)} & \textbf{Precision (\%)} & \textbf{Recall (\%)} & \textbf{Training Time (h)} \\ \hline
\multicolumn{7}{c}{\cellcolor[HTML]{E2E0E0}\textbf{VGG16 (approach 1)}} \\ \hline
DIF & False & \textbf{97.02} & \textbf{96.80} & 97.38 & \textbf{96.23} & 4.45 \\
DIF & True & 94.76 & 95.02 & 95.66 & 94.37 & 4.47 \\
ADD & False & 95.50 & 96.28 & \textbf{97.79} & 94.82 & \textbf{4.44} \\
ADD & True & 94.48 & 94.53 & 94.99 & 94.03 & 4.54 \\ \hline
\multicolumn{7}{c}{\cellcolor[HTML]{E2E0E0}\textbf{ResNet-50 (approach 1)}} \\ \hline
DIF & False & \textbf{98.42} & \textbf{98.27} & \textbf{98.52} & \textbf{98.03} & 4.45 \\
DIF & True & 94.37 & 94.34 & 95.24 & 93.47 & 4.47 \\
ADD & False & 96.34 & 96.26 & 96.65 & 95.83 & \textbf{4.44} \\
ADD & True & 94.87 & 95.76 & 97.05 & 94.48 & 4.46 \\ \hline
\multicolumn{7}{c}{\cellcolor[HTML]{E2E0E0}\textbf{ResNet-50 with attention (approach 2)}} \\ \hline
DIF & False & 99.04 & 99.03 & 99.04 & 99.04 & 2.93 \\
DIF & True & 98.31 & 98.32 & 98.37 & 98.25 & 2.76 \\
ADD & False & 99.38 & 99.39 & 99.38 & 99.38 & 3.86 \\
ADD & True & \textbf{99.61} & \textbf{99.61} & \textbf{99.61} & \textbf{99.61} & 4.17 \\ \hline
\end{tabular}%
}
\end{table}

Table \ref{tab:grad_cams_results} shows the results obtained when evaluating each model's grad-CAMs. Contrarily to the quantitative classification results for VGG16 and ResNet-50, the cropped inputs presented better focus, meaning a higher similarity between the model's grad-CAMs and the GT masks. In average, cropping part of the background from the inputs increased the Dice and IoU metrics by 4.72\% and 3.16\%, for the \emph{DIF} input, and by 9.09\% and 5.98\%, for the \emph{ADD} input, respectively. The cropped \emph{ADD} led to a better focus, achieving its best results with the ResNet-50. This model showed an overall average boost of 7.03\% in Dice and 4.96\% in IoU metric, when compared to the VGG16. 

The attention mechanism improved the focus of the ResNet-50 for every input, even if just for a small margin (<5\%). The greatest improvement was achieved by the non-cropped \emph{ADD} input (4.20\% in Dice). However, this one still led to lower metrics than the non-cropped \emph{DIF} input, as the latter achieved Dice values 4.07\% higher than the non-cropped \emph{ADD}. Contrarily to previous results, cropped \emph{DIF} attained the highest similarity between their GT masks and the grad-CAMs obtained from the ResNet-50 model with attention. Nonetheless, the difference relative to the cropped \emph{ADD} corresponds only to a small percentage (1.11\%).  Additionally, the cropped \emph{ADD} presented the lowest values of standard deviation, entailing a more consistent focus across the dataset, with smaller fluctuations and more representative IoU and Dice values.

\begin{table}[H]
\centering
\caption{Quantative evaluation results, in percentage, for validation grad-CAMs, when predicting with the VGG16 and ResNet-50 models (without and with an attention mechanism)}
\label{tab:grad_cams_results}
\maxsizebox{0.85\textwidth}{!}{%
\begin{tabular}{@{}cc|cc|cc|cc@{}}
\toprule
\multicolumn{2}{c|}{\textbf{Input}} & \multicolumn{2}{c|}{\textbf{VGG16}} & \multicolumn{2}{c|}{\textbf{ResNet-50}} & \multicolumn{2}{c}{\textbf{ResNet-50 with attention}} \\ \midrule
\textbf{Type} & \textbf{Crop} & \textbf{\begin{tabular}[c]{@{}c@{}}Dice ($\pm{std}$)\end{tabular}} & \textbf{\begin{tabular}[c]{@{}c@{}}IoU ($\pm{std}$)\end{tabular}} & \textbf{\begin{tabular}[c]{@{}c@{}}Dice ($\pm{std}$)\end{tabular}} & \textbf{\begin{tabular}[c]{@{}c@{}}IoU ($\pm{std}$)\end{tabular}} & \textbf{\begin{tabular}[c]{@{}c@{}}Dice ($\pm{std}$)\end{tabular}} & \textbf{\begin{tabular}[c]{@{}c@{}}IoU ($\pm{std}$)\end{tabular}} \\
DIF & False & 16.16 ($\pm{9.64}$) & 9.10 ($\pm{5.87}$) & 29.06 ($\pm{13.26}$) & 17.70 ($\pm{9.01}$) & 30.37 ($\pm{12.86}$) & 18.59 ($\pm{9.13}$) \\
DIF & True & 22.57 ($\pm{14.26}$) & 13.44 ($\pm{8.95}$) & 32.09 ($\pm{11.57}$) & 19.67 ($\pm{8.22}$) & \textbf{32.90} ($\pm{9.13}$) & \textbf{20.04} ($\pm{6.43}$)\\
ADD & False & 20.19 ($\pm{6.84}$) & 11.39 ($\pm{4.21}$) & 22.10 ($\pm{16.00}$) & 13.36 ($\pm{10.51}$) & 26.30 ($\pm{14.71}$) & 15.99 ($\pm{10.04}$) \\
ADD & True & \textbf{28.34} ($\pm{9.17}$) & \textbf{16.84} ($\pm{6.26}$) & \textbf{32.13} ($\pm{13.34}$) & \textbf{19.89} ($\pm{9.48}$) & 31.79 ($\pm{8.54}$) & 19.21 ($\pm{6.13}$)\\ \bottomrule
\end{tabular}%
}
\end{table}

\subsection{Segmentation-Classification Approach}
\label{sec:results_segmentclassif}

Table \ref{tab:seg_results} shows the validation results obtained for segmentation, along with the training time. The segmentation and following classification training curves can be found in \ref{appendix_seg}. Considering these results, all inputs seem to accurately segment the human body region, with Dice and IoU values higher then 93\%. Among these inputs, the cropped \emph{ADD} stood out, although for a small margin (less than 1\%), presenting an IoU and Dice of 94.52\% and 97.17\%, respectively. The segmentation test results, along with examples of segmented images can be found in \ref{appendix_seg}.

\begin{table}[H]
\centering
\caption{Validation results, in percentage, of the UNET model, as well as the training time for 30 epochs}
\label{tab:seg_results}
\maxsizebox{0.85\textwidth}{!}{%
\begin{tabular}{ccccccc}
\hline
\textbf{Input Type} & \textbf{Crop} & \textbf{Accuracy} & \textbf{Loss} & \textbf{IoU ($\pm{std}$)} & \textbf{Dice ($\pm{std}$)} & \textbf{Training Time (h)} \\ \hline
DIF & False & 98.80 & 0.02 & 93.81 ($\pm{1.96}$) & 96.80 ($\pm{1.05}$) & 1.37 \\
DIF & True & 98.36 & 0.03 & 93.85 ($\pm{1.95}$) & 96.81 ($\pm{1.05}$) & 1.37 \\
ADD & False & 98.94 & 0.02 & 94.34 ($\pm{1.71}$) & 97.08 ($\pm{0.91}$) & 1.38 \\
ADD & True & 98.57 & 0.03 & \textbf{94.52} ($\pm{1.71}$) & \textbf{97.17} ($\pm{0.91}$) & 1.37 \\ \hline
\end{tabular}%
}
\end{table}

Tables \ref{tab:segclass_val_results} and \ref{tab:segclass_cams_results} present the validation results for classification, as well as the training time for 100 epochs, and the quantitative evaluation of the model focus, respectively. The \emph{DIF} input attained the highest values of F1-score, namely 94.14\% and 93.36\% for non-cropped and cropped, respectively. However and once again, the best focus was achieved with the cropped \emph{ADD} input (Dice values with average of 28.17\%).

\begin{table}[H]
\centering
\caption{Validation results of the adapted UNET classification model, following the segmentation task, as well as the training time for 100 epochs}
\label{tab:segclass_val_results}
\resizebox{0.85\textwidth}{!}{%
\begin{tabular}{@{}cccccccc@{}}
\toprule
\textbf{Input Type} & \textbf{Crop} & \textbf{ACC (\%)} & \textbf{Loss} & \textbf{F1-score} & \textbf{Precision (\%)} & \textbf{Recall (\%)} & \textbf{Training Time (h)} \\ \midrule
DIF & False & \textbf{94.09} & \textbf{0.16 }& \textbf{94.14} & \textbf{94.29} & \textbf{93.92} & 4.48 \\
DIF & True & 93.47 & 0.17 & 93.36 & 93.70 & 93.02 & 4.53 \\
ADD & False & 90.82 & 0.24 & 91.08 & 92.07 & 90.23 & 4.49 \\
ADD & True & 92.79 & 0.27 & 92.69 & 93.08 & 92.34 & \textbf{4.43} \\ \bottomrule
\end{tabular}%
}
\end{table}

\begin{table}[H]
\centering
\caption{Quantitative focus evaluation results, in percentage, of the validation grad-CAMs, when predicting with the adapted UNET model for classification}
\label{tab:segclass_cams_results}
\maxsizebox{0.4\textwidth}{!}{%
\begin{tabular}{@{}cccc@{}}
\toprule
\textbf{Input Type} & \textbf{Crop} & \textbf{Dice ($\pm{std}$)} & \textbf{IoU ($\pm{std}$)} \\ \midrule
DIF & False & 19.48 ($\pm{8.00}$) & 11.01 ($\pm{5.06}$) \\
DIF & True & 18.27 ($\pm{10.46}$) & 10.44 ($\pm{6.80}$) \\
ADD & False & 21.21 ($\pm{7.80}$) & 12.08 ($\pm{4.88}$) \\
ADD & True & \textbf{28.17} ($\pm{9.66}$) & \textbf{16.75} ($\pm{6.35}$) \\ \bottomrule
\end{tabular}%
}
\end{table}

\subsection{Analysis of the final approach}
\label{sec:results_segmentclassif}

The best classification performance, as well as the most relevant and human-centred focus, were achieved by the ResNet-50 model with a channel-wise attention mechanism (\textbf{approach 2}). Therefore, further evaluation is conducted to analyse this model's performance, in both offline and trial simulations, with an unseen test dataset.

\subsubsection{Offline testing}
\label{sec:offline_test}

Table \ref{tab:att_test_results} presents the percentage of wrong classifications over the test set. Similarly to validation, the cropped \emph{ADD} revealed the most promising results, with the wrong predictions representing only 0.64\% of the dataset.
    
\begin{table}[H]
\centering
\caption{Percentage of wrongly classified frames in the test set, by the ResNet-50 model with an attention mechanism}
\label{tab:att_test_results}
\maxsizebox{0.4\textwidth}{!}{%
\begin{tabular}{@{}ccc@{}}
\toprule
\textbf{Input Type} & \textbf{Crop} & \textbf{Wrong predictions (\%)} \\ \midrule
DIF & False & 0.68 \\
DIF & True & 2.38 \\
ADD & False & 0.71 \\
ADD & True & \textbf{0.64} \\ \bottomrule
\end{tabular}%
}
\end{table}

\begin{figure}[t]
    \centering
    \includegraphics[width=0.7\textwidth]{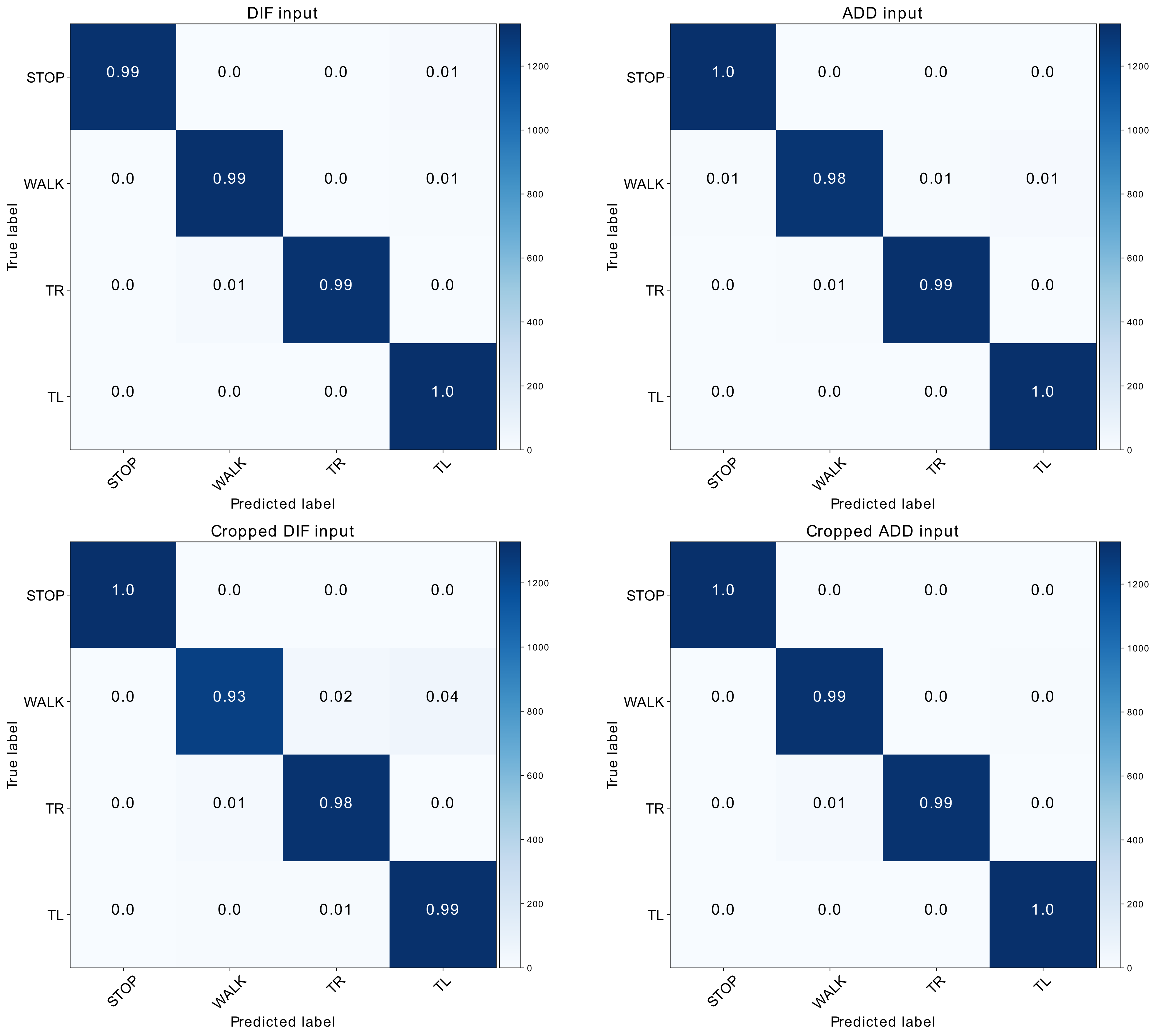}
    \caption{Confusion matrices for the ResNet-50 model with attention, over the 4 forms of input.}
    \label{fig:att_test_results}
\end{figure}

Figure \ref{fig:att_test_results} presents the obtained confusion matrices. These matrices reveal excellent results, specially for the stop and TL classes, where the \emph{TP} rate was never lower than $0.99$. The cropped \emph{DIF} was the least favoured input by this architecture, with the lowest \emph{TP} rate for the walk class ($0.93$).

Table \ref{tab:att_cams_results_appendix} shows the results obtained from the focus evaluation of ResNet-50 model with attention. The input influence over the model focus is in agreement with the one already verified in validation (Table \ref{tab:grad_cams_results}). Cropped \emph{DIF} attained the higher similarity between their GT masks and the grad-CAMs, but with a very small difference from the cropped \emph{ADD} results (without surpassing 0.16\%, considering both Dice and IoU). The standard-deviation obtained for the cropped \emph{ADD} was also smaller in comparison with that obtained for the cropped \emph{DIF}, similar to what was found for the validation set.

\begin{table}[H]
\centering
\caption{Quantitative focus evaluation results, in percentage, of the test grad-CAMs, when predicting with the ResNet-50 model with an attention mechanism}
\label{tab:att_cams_results_appendix}
\maxsizebox{0.4\textwidth}{!}{%
\begin{tabular}{@{}cccc@{}}
\toprule
\textbf{Input Type} & \textbf{Crop} & \textbf{Dice ($\pm{std}$)} & \textbf{IoU ($\pm{std}$)}\\
\midrule
DIF & False & 29.97 ($\pm{14.56}$) & 18.50 ($\pm{10.26}$) \\
DIF & True & \textbf{32.38} ($\pm{10.32}$) & \textbf{19.76} ($\pm{7.20}$) \\
ADD & False & 22.14 ($\pm{13.02}$) & 13.07 ($\pm{8.59}$) \\
ADD & True & 32.30 ($\pm{8.83}$) & 19.60 ($\pm{6.37}$) \\ \bottomrule
\end{tabular}%
}
\end{table}

\subsubsection{Trial simulations}
\label{sec:realtime_simulations}

The cropped \emph{ADD}, when fed to the ResNet-50 model with attention, achieved the best results across validation and test, considering both classification power and focus evaluation. Therefore, this was the approach tested in trial simulations, along with the post-processing technique. Two representative trials, corresponding to different conditions and extreme cases, are displayed: \textbf{trial A)} participant 11 performs a TL at 0.5m/s (lowest gait speed); \textbf{trial B)} participant 15 performs a TR at 1m/s (fastest gait speed).

\begin{figure}[t]
    \centering
    \includegraphics[width=0.7\textwidth]{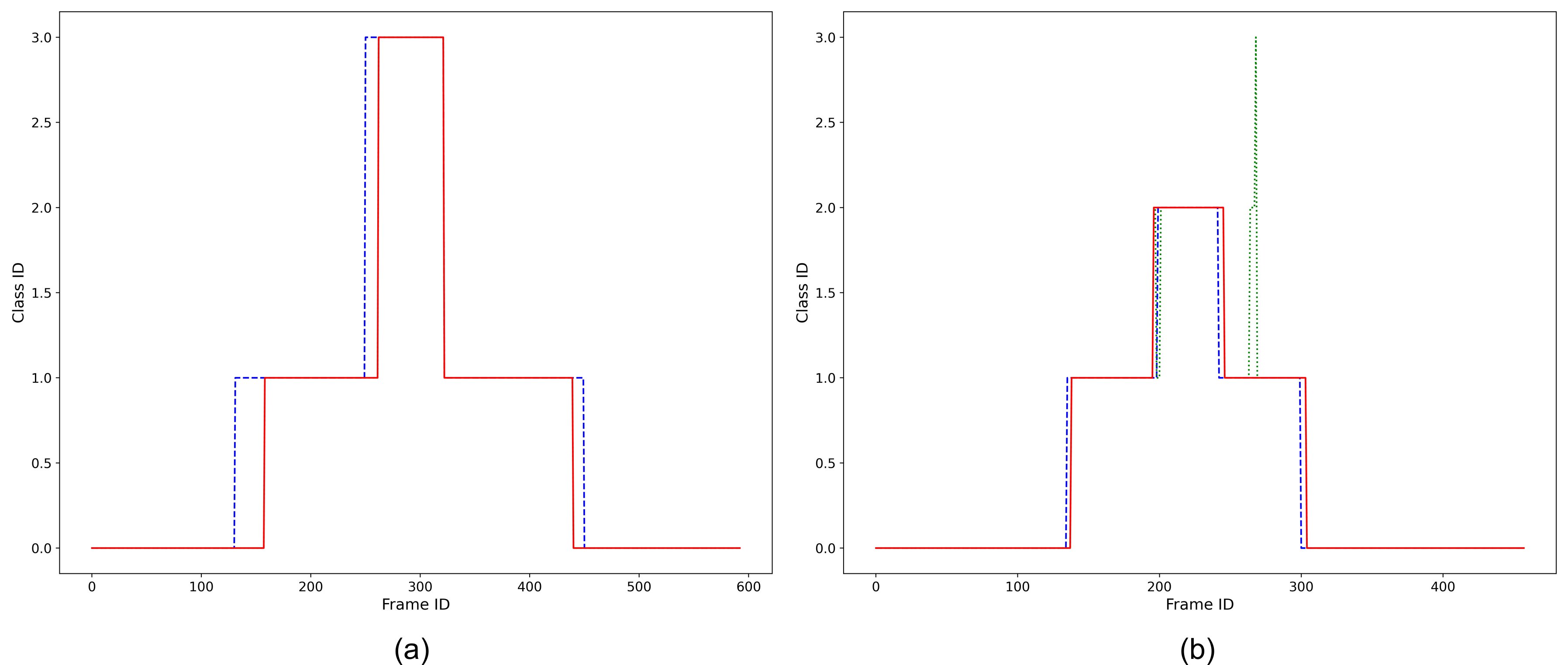}
    \caption{Plot of the GT (dashed line), predicted (dotted line) and post-processed predicted labels (solid line) for (a) Trial A and (b) Trial B. Class IDs: 0=stop, 1=walk, 2=TR, 3=TL.}
    \label{fig:realtime_predictions}
\end{figure}

\begin{figure}[t]
    \centering
    \includegraphics[width=0.7\textwidth]{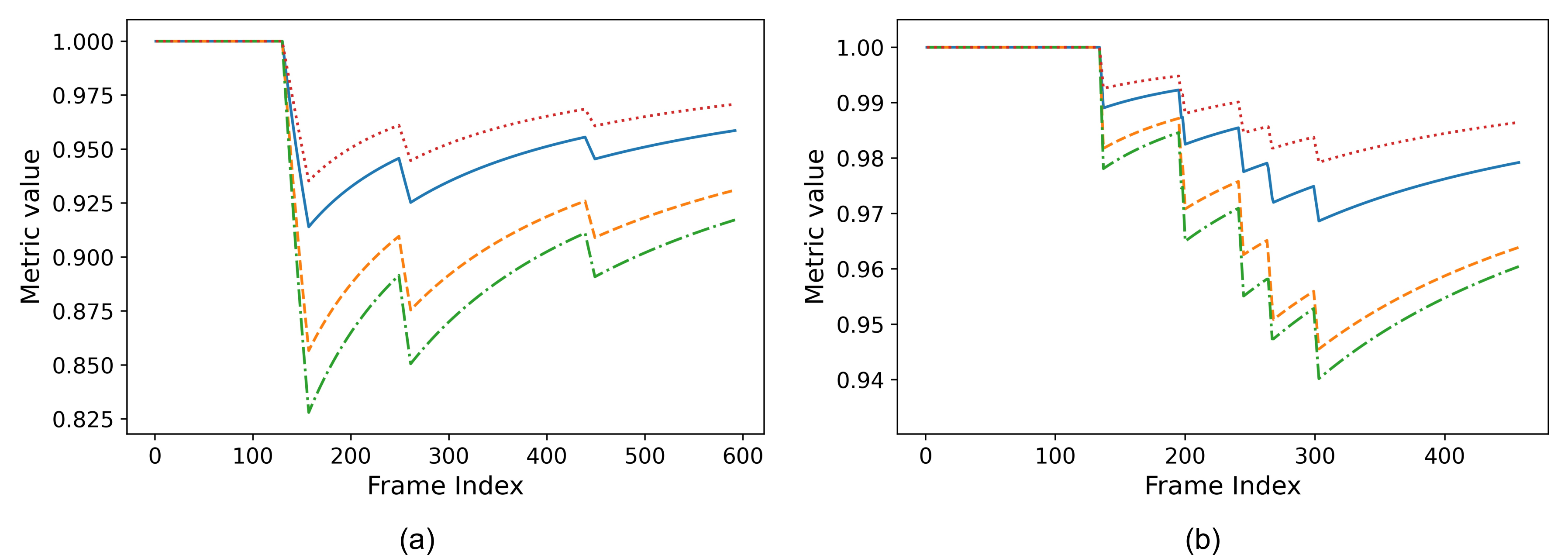}
    \caption{Plot of the values of the online metrics described in Section \ref{sec:metrics} for (a) Trial A and (b) Trial B, namely: instantaneous accuracy (IA, solid line), instantaneous weighted accuracy (wIA, dashed line), instantaneous precision (IP, dashed line with dots) and instantaneous calibrated precision (cIP, dotted line).}
    \label{fig:realtime_metrics}
\end{figure}

Figure \ref{fig:realtime_predictions} allows the comparison, at each instant, between the predictions (with and without post-processing) and the GT classes. Figure \ref{fig:realtime_metrics} shows the temporal evolution of the online metrics described in Section \ref{sec:metrics}.

Table \ref{tab:realtime_delays} shows, respecting the trials' temporal order, the delays between the post-processed label and the GT one, for each action transition. Negative values represent classes predicted earlier than their actual start.

\begin{table}[H]
\centering
\caption{Delays of the final predicted labels in relation to the respective GT labels, computed for each transition of the circuit}
\label{tab:realtime_delays}
\maxsizebox{0.4\textwidth}{!}{%
\begin{tabular}{@{}ccccc@{}}
\toprule
\multicolumn{5}{c}{\textbf{Time delay (s)}} \\ \midrule
\textbf{Trial} & \textbf{Walk} & \textbf{Turn} & \textbf{Walk} & \textbf{Stop} \\ \midrule
\textbf{A} & 1.80 & 0.80 & 0.00 & -0.67 \\
\textbf{B} & 0.20 & -0.20 & 0.27 & 0.27 \\ \bottomrule
\end{tabular}%
}
\end{table}

\subsubsection{Grad-CAMs visualisation}
\label{sec:results_realtime_cams}

\begin{figure}[t!]
    \centering
    \includegraphics[width=0.7\textwidth]{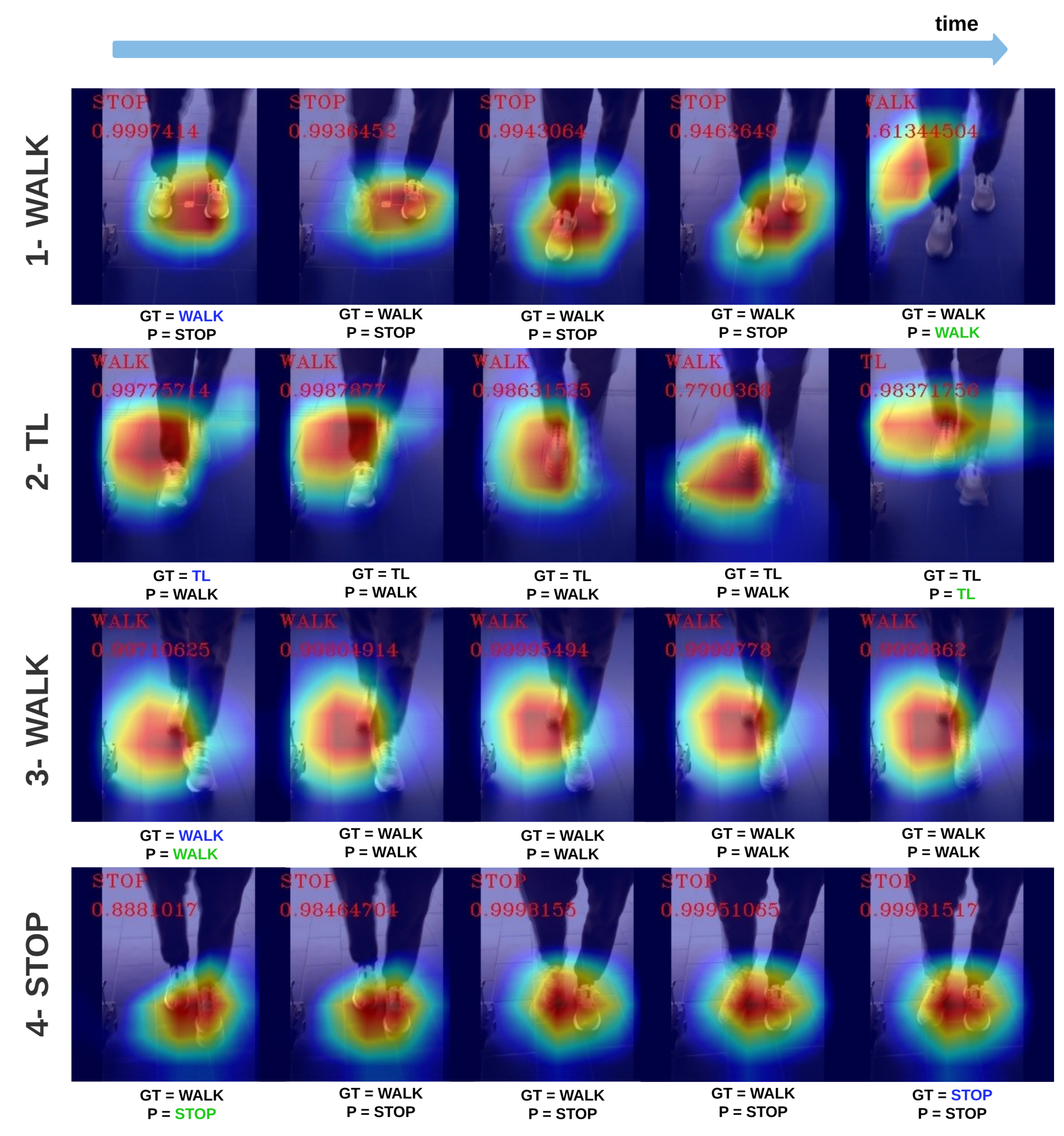}
    \caption{Grad-CAMs visualisation, temporally ordered, for each one of the transitions in the slow trial (\textbf{trial A}). The green and blue labels correspond to the first prediction and GT label, respectively, of the action that is beginning (P=predicted class).}
    \label{fig:slowtrial_cams}
\end{figure}

Figure \ref{fig:slowtrial_cams} presents the grad-CAMs visualisation for \textbf{Trial A}, where the model's predicted labels correspond exactly to the post-processed ones. For the beginning of each class, the visualisation starts at the first frame of that action (for delayed predictions) or at the first correct prediction (for early predictions, as it is the case of the stop class, in this trial) and ends at the first right prediction or first GT frame, respectively. Note that these are not necessarily consecutive frames on the dataset, that depends on the action delay registered in Table \ref{tab:realtime_delays}. Nonetheless, they serve as a good representation of the focus evolution between the GT and its respective correct prediction (or vice-versa) and, for the delayed predicted labels, it always includes two immediately preceding frames.

\begin{figure}[t!]
    \centering
    \includegraphics[width=0.7\textwidth]{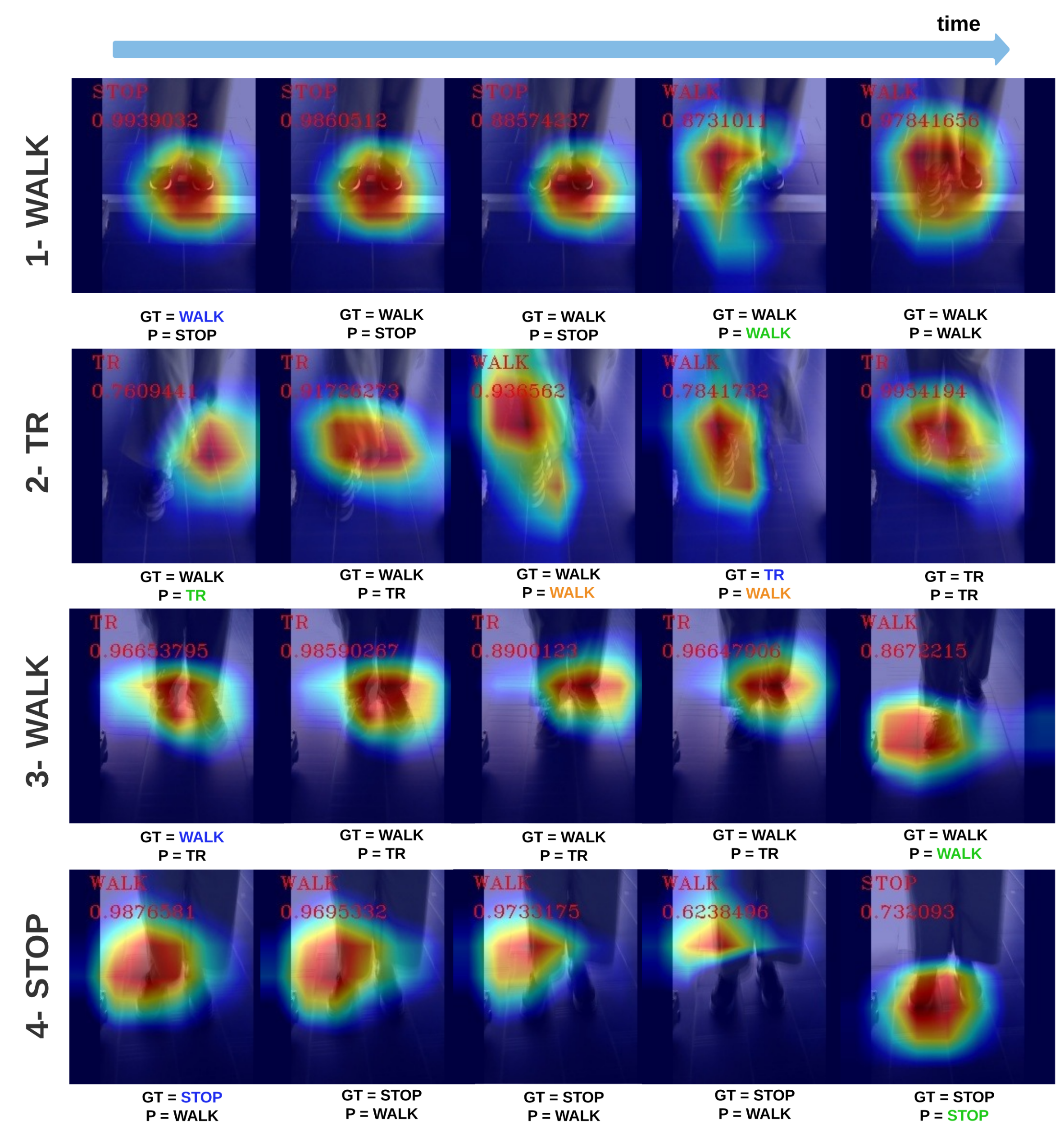}
    \caption{Grad-CAMs visualisation, temporally ordered, for each one of the transitions in the fast trial (\textbf{trial B}). The green and blue labels correspond to the first prediction and GT label, respectively, of the action that is beginning (P=predicted class). The orange ones correspond to the perturbations in the model's predictions that do not correspond to the post-processed predicted class.}
    \label{fig:fasttrial_cams}
\end{figure}

The same applies to Figure \ref{fig:fasttrial_cams}, representing \textbf{Trial B}. This trial presents more on-off noise in the model's outcomes, specially in the transition from walk to turn (Figure \ref{fig:realtime_predictions}b). Hence, its predictions do not always correspond to the post-processed labels. As the latter constitutes the final predicted classes, the beginning or end of these visualisations are depicted by the respective post-processed labels, for early and late predictions, respectively. Moreover, in the two first presented classes, a frame immediately after the correct post-processed prediction/GT label was added for purposes of focus evolution assessment.

\section{Discussion}
\label{sec:discussion}

The results obtained by the DL approaches are critically discussed in this section, pointing some limitations and insights for possible improvements.

\subsection{Tailored inputs and model's focus evaluation}
\label{sec:discussion_inputs}

Distinct input forms influence the models performance differently. Despite leading, in most cases, to lower classification metrics, cropping the images helps to direct the focus to the human ROI (see Table \ref{tab:grad_cams_results}), as it excludes a significant portion of background. Therefore, better classification results can be associated with less reliable extracted features. 

A greater improvement in focus was registered when cropping the \emph{ADD} input, which confirms that non-cropped \emph{ADD} includes more information about the background motion. The fact that the highest improvement rate in focus achieved by adding an attention mechanism to the ResNet-50 model was recorded by the non-cropped \emph{ADD} also supports this statement. According to the results, the cropped \emph{ADD} was the input that consistently presented high similarities between grad-CAMs and GT masks, raising the belief that \emph{ADD} images also encode more human body motion information. 

From these comparisons, three aspects can be inferred: \textbf{i)} background motion may contain more evident features, easier to extract, that can help the offline classification task. However, this should not be the main focus of the model since these features are not reliable for detecting transitions, real-time applications or even generalisations to other datasets, where the background is static; \textbf{ii)} a careful performance evaluation is needed, since better classification results can be associated with non-ideal feature extractions and overfitting to the background; and \textbf{iii)} despite not being a commonly used approach, cropping the inputs was a feasible way to enhance the model focus and thus increased the reliability of the respective classification results. Nevertheless, the higher registered improvement was not higher than 10\% (see Table \ref{tab:grad_cams_results}). This can be related to the background characteristics, such as the presence of floor stripes enhancing background motion (Figures \ref{fig:slowtrial_cams} and \ref{fig:fasttrial_cams}).

Literature on HAR/HAP normally uses sequences of RGB frames to provide the sense of temporal motion. Instead of this approach, the proposed inputs were able to provide (a significant part of) this temporal information in one single frame, avoiding the use of Recurrent Neural Networks. With a larger background variability in the dataset, while carefully avoiding pavement marks during the acquisitions, these tailored inputs could become a more reliable method to induce action-aware feature extraction, avoiding more efficiently the background features.

Nevertheless, the lower metrics obtained when evaluating the model focus can also be due to the quantitative focus evaluation algorithm itself. The GT masks used here as a comparison term are binary, presenting the highest score (1) for all the human area, but this region may not be equally important to decode human motion. For example, feet and knees may present more orientation and position variations which indicate the step's direction, so it would be correct for the model to give higher focus to these particular regions. For this reason, comparing heatmaps to these masks is correctly penalising \emph{FP}, but it is also punishing the model for not focusing on the complete lower body, including, for example, the more static upper leg region. These effects can lower the values of the calculated Dice and IoU. The masks are already computed in the tightest ROI possible to attenuate this effect, but it does not completely solve it. A possible solution for this problem would be to change the GT masks pixel values, according to the input images. Thus, as the higher pixel intensities in the input correspond to motions with larger amplitudes, while the lower correspond to more static areas, this information could be used to scale the scores equal to 1 in the GT masks, creating a sort of \emph{human motion masks}. An even more accurate form of information to scale these masks according to the body's amplitude of motion, would be the human poses, from the Xsens data, for instance. Nevertheless, this last option would unduly increase the computational expense and complexity of this algorithm. Observing the grad-CAMs of the final selected algorithm (Figures \ref{fig:slowtrial_cams} and \ref{fig:fasttrial_cams}), one could also attempt to only generate masks of the user's feet.

\subsection{Single-frame Classification Approach }
\label{sec:results_segmentclassif}

Based on the results presented in Table \ref{tab:input_val_results}, it is possible to notice that ResNet-50 performed better than VGG16, achieving higher F1-score values than VGG16 ([94.34\%, 98.27\%] over [94.53\%, 96.80\%], respectively). Hence, this problem benefited from residual features, skip connections and deeper networks duly initialised or pre-trained. Nevertheless, the difference between both performances was not that high (lower than 1.47\% in F1-score - Table \ref{tab:input_val_results}), meaning that the task of early recognising human motions from the SW dataset may also be approached by less deep models. Nonetheless, ResNet-50 achieved better classification results and focused better on the human region (average boost of 7.03\% in Dice and 4.96\% in IoU), over all input forms, and was, thus, the chosen model to be tested with an attention mechanism (\textbf{approach 2}).

The addition of the channel-wise attention mechanism enhanced not only the classification metrics, improving the F1-score by an average of 2.93\%, but also the similarity between grad-CAMs and GT masks, with improvements until 4.20\% in Dice, across all inputs. Only the model focus associated with the cropped \emph{ADD} input was slightly worse than the ones registered with the ResNet-50 baseline model (see Table \ref{tab:grad_cams_results}). However, since the difference is not that large (0.34\% and 0.68\% in Dice and IoU, respectively), this could be due to small variations and, thus, was not considered as a relevant fact.

These results proved the importance of dealing and modelling the distinct learning abilities of the different convolutional channels, not only to increase CNN performance, but also to improve the relevance of the features extracted. However, the values presented in Table \ref{tab:grad_cams_results} for ResNet-50 with attention are not that higher than the ones for the corresponding baseline model, specially for the cropped inputs, proving that this channel-wise attention mechanism, although unequivocally beneficial to the classification task, still does not completely correct its main focus, as Dice and IoU are still below 50\% and most pixels in grad-CAMs heatmaps do not correspond to the human body region. 

Facing these facts, a spatial attention mechanism could also be designed for this problem, guiding the model to use \textit{attentional regions}, instead of the whole frame. As in \citet{Wu2020}, also enhancing local features by combining this with the channel-wise mechanism, could lead to better performances and, in this case, more properly focused solutions. The spatial attention maps could even be compared with the suggested \emph{human motion masks} for focus evaluation or, in a more bold experiment, as part of the training loss.

The cropped \emph{ADD} appeared as the most promising input for this task and, when fed to the ResNet-50 model with the attention mechanism, achieved the best overall results.

\subsection{Segmentation-Classification Approach }
\label{sec:results_segmentclassif}

Looking at the segmentation training curves (Figure \ref{fig:seg_trainplots}), one can see that, as the epochs advance, there is a tendency for overfitting, given the small increase in the gap between the validation the training losses. Although apparently small, this can propagate to the following pre-trained classification model and induce bad generalisation abilities or even worse cases of overfitting. That is why the segmentation training was shorten to 30 epochs and the weights were chosen considering the minimal validation loss. Despite the training reduction, the adapted UNET still revealed problems of weak generalisation (Figure \ref{fig:segclass_trainplots}). In agreement with these training curves, the classification metrics were worse than the ones achieved by previous evaluated models, as the maximum F1-score was of 94.14\% (Table \ref{tab:segclass_val_results}), which is lower than the minimum registered for the previous models (94.34\%, for the baseline ResNet-50, shown in Table \ref{tab:input_val_results}). Nevertheless, these metrics were still above 90\%. The severest cases of unrepresentative training dataset and consequent generalisation issues were verified by the \emph{ADD} input type, which is associated with lower validation performances, namely 92.69\% (cropped) and 91.08\% (non-cropped form) of F1-score (Table \ref{tab:segclass_val_results}).

When connecting the segmentation and classification results (Tables \ref{tab:seg_results} and \ref{tab:segclass_val_results}, along with Figure \ref{fig:segclass_trainplots}), it seems to exist an inverse relation between segmentation power and the classification generalisation ability. This may mean that this cascade approach is leading the model to focus on input traits that are not representative of the whole dataset, following the overfitting problems during segmentation. Cross-validation should be performed to prove this statement. Even so, the focus on particular traits can be associated with the fact that, despite the final aim of human motion decoding, the GT masks used are leading to the segmentation of the whole body, including large clothes and more static human areas. Therefore, using the aforementioned \emph{human motion masks} as labels could decrease the chances of overfitting, while pursuing the differential segmentation of the human body, according to its motion. This could enhance the weights used to pre-train the classification model. Other options to help overcoming the overfitting problem consist on experimenting other simpler segmentation models or even include spatial data augmentation. Moreover, the number of frozen layers should also be studied and tuned.

As for the grad-CAMs evaluation (Table \ref{tab:segclass_cams_results}), the segmentation-classification approach was not the most effective to attain its main goal: improving the extraction of human-centred features during classification. The low results, along with their resemblance to the ones achieved by VGG16, point to an influence mainly exerted by the input properties and not by the two-stage framework itself.

\subsection{Trial simulations}
\label{sec:results_segmentclassif}

The ResNet-50 model with a channel wise attention mechanism was the best model in both aspects: classification rates and focus relevance, specially when fed with the cropped \emph{ADD} input. Testing this approach in trial simulations led to general good performances, with the model being able to identify the different consecutive walking events. 

The model uncertainty revealed a greater prominence at higher gait speeds, in the form of on-off noise (see Figure \ref{fig:realtime_predictions}). However, these perturbations were easily smoothed by the proposed post-processing technique, without adding time delays in the correct predictions. Despite presenting more noise, \textbf{Trial B} achieved overall higher online metric values, over 94\%, due to its lower time delays registered in transitions.

Table \ref{tab:realtime_delays} shows, respecting the trials' temporal order, the delays between the post-processed and the GT labels, for each action transition. For the trial performed at 1 m/s, the delays are at least 0.37s lower than the average step time for this gait speed (0.64s). For lower gait speeds, the time lags registered were higher, confirming the greater challenge implied by early detecting slower and more subtle motion changes. These delays could be further decreased for real-time applications, through a proper training procedure that includes transitions in the dataset, but also through a data quality enhancement to further improve the model focus (\emph{e.g.} large variety of backgrounds without floor stripes/marks).

Overall, the results prove the chosen approach is suitable for early action recognition, achieving average online metrics between 91.72\% and 98.65\%. However, it still needs improvements to early detect an action, as it can be seen by the model performances at the transition inputs. Nevertheless, the obtained results were still good, with not so critical delays, considering the fact that the neural network was not trained with transitions. Hence, with the mentioned suggestions, specially the inclusion of transitional frames in the training procedure, this performance could be further enhanced. For this to happen, one has to first improve the labelling accuracy, decreasing the bias effect introduced by the person controlling the joystick.

\subsubsection{Grad-CAMs visualisation}
\label{sec:results_realtime_cams}

The displayed grad-CAMs (see Figures \ref{fig:slowtrial_cams} and \ref{fig:fasttrial_cams}) showed that the model focus is not too deviated from the human region, but it still considers some background information, specially the visible motion of the floor stripes. The floor stripes were a non-ideal property of the acquisition environment, which visibly affected the model training. As background motion is a consequence of the walker's movement, and not the user's, this misleading focus can be among the causes of the registered time delays, when predicting each action's beginning.

For example, in Figure \ref{fig:fasttrial_cams} (last row), the stop detection was delayed, as the walker kept moving after the subject stopped. So the model must have considered the stripes and the large clothes motions, instead of the user's steadier positions. As the device decelerates, this background motion became less evident and the model started to focus on the feet. In the slow trial, this deceleration phase is shorter and slower, so the background motion stopped appearing in the RGB inputs before the human movement, allowing the model to better perceive the progressive horizontal alignment of the feet (last row of Figure \ref{fig:slowtrial_cams}). This situation is similar to the beginning of the first walking event at low gait speed (first row), where the walker starts to slowly accelerate, so the background appears static, and the feet move slowly and closer to each other, leading to a confusion between stop and walk classes. It seems the model cannot yet associate the horizontally misalignment of the feet as a walking trait.

The turning event was anticipated in \textbf{Trial B}, which seemed like a good achievement for motion intention decoding. Nonetheless, looking at the respective grad-CAMs (second row, in Figure \ref{fig:fasttrial_cams}), one can see that this class was first predicted based on the vertical misalignment between the floor stripes. This helps to visualise and understand the confusion and model uncertainty between these two classes (walk and TR/TL). 

These visualisations showed that the model focus still needs to be improved in transitions, in order to be integrated in a real-time control mechanism. Nevertheless, this focus appeared to be better than the expected from its quantitative evaluation, with grad-CAMs concentrated around the feet, which can be due to the GT masks used in that algorithm.

\section{Conclusion}
\label{sec:conclusions}

This work presents a novel way to decode motion intention in SWs. Three different approaches were devised, two facing single-frame classification, and another facing a segmentation-classification approach. For that, a custom dataset of 15 healthy participants was acquired with a smart robotic walker, considering realistic scenarios and circuits. Each participant performed a total of 24 trials, each one containing three of the target classes (stop, walk, turn right and turn left). Considering this, a balanced dataset of frames containing 28800 RGB-D images was created, extracting 40 frames per video sequence, and used to train, evaluate, and compare the proposed approaches.

Regarding the model architectures, state-of-the-art VGG16 and ResNet-50 were implemented in the first approach and then, an attention mechanism was added to the best model (second approach). For the third approach, the UNET neural network was used for segmentation and adapted for the following classification task. Four different input forms were studied, namely cropped and non-cropped \emph{ADD} and \emph{DIF} images. These were obtained considering a sliding window approach of 4 frames with a stride of 2, by summing the four frames of the window (\emph{ADD}) or subtracting the last frame from the first (\emph{DIF}). This approach encoded 0.27s of motion information without using recurrent neural networks, which are a commonly used approach in the literature. To evaluate the model performance, we considered standard metrics, namely accuracy, F1-score, recall, and precision. For trial simulations assessments, OAD metrics were used, such as instantaneous accuracy, instantaneous precision, instantaneous weighted accuracy, and instantaneous calibrated precision. We also evaluated the model focus considering a novel method that quantitatively compares its grad-CAMs with GT masks of the human body region. 

Regarding the different inputs, we concluded that the non-cropped \emph{ADD} input encodes more motion information, but these, together with \emph{DIF} images share a common disadvantage with the optical flow: in realistic videos, these inputs also encode background motion which may deviate the model focus from the human region. Cropping most part of the background surrounding the image's ROI proved to have a major impact on the model's focus. We also verified that ResNet-50 with an attention mechanism, when fed with cropped \emph{ADD} inputs, attained the most promising results (offline accuracy and F1-score higher than 95\%). This enabled an enhancement in the model focus towards the human body region (Dice rounding 32\%) when compared to the other models, but still needs further improvements. 

\subsection{Limitations and future research insights}

This work presents some limitations that should be considered in future research insights. Enhancing the quality of the acquired data becomes necessary to train the algorithms with transitional inputs and improve the relevance of the extracted features. Recording in a more controlled environment, without marked floors or too bright conditions, would be relevant to not deviate the model focus to the background. Moreover, the labelling procedure should also be improved (\textit{e.g.} with the use of force sensors) to allow the inclusion of transitions, without mislabelled samples. Data from pathological individuals should be acquired and the use of transfer learning may be considered to endow the model with the ability to detect motion intentions for this population. Regarding methodology, the method to evaluate the model focus can be refined. For instance, the human masks can be improved by lowering the pixel values in more static human body areas, creating \emph{human motion masks}, along with the inclusion of false positives as a metric to accurately assess how much the model is focusing on the background. Spatial attention mechanisms or self-supervised learning can also be explored to improve the model focus. Additionally, tailored losses could also be tested to improve the early detection ability, especially if the model keeps presenting significant time delays. Lastly, the use of visual transformers could also be investigated to progressively obtain simpler models capable of learning by observation. 

Along with its potential for improvement, we hope this work can also serve as future benchmark and encourage further investigations on decoding fine-grained human actions directly through visual information.

\section*{Acknowledgements}
This work has been supported by the \emph{Fundação para a Ciência e Tecnologia} (FCT) with the Reference Scholarship under Grant 2020.05708.BD and under the national support to R\&D units grant, through the reference project UIDB/04436/2020 and UIDP/04436/2020.

\section*{Author Contributions}
Carolina Gonçalves: Methodology, Investigation, Data curation, Formal analysis, Software, Writing - Original Draft, Writing - Review \& Editing; João M. Lopes: Methodology, Investigation, Data curation, Software, Writing - Review \& Editing, Funding acquisition; Sara Moccia: Methodology, Resources, Supervision, Validation, Writing - Review \& Editing; Daniele Berardini: Software, Writing - Review \& Editing; Lucia Migliorelli: Software, Writing - Review \& Editing; Cristina P. Santos: Conceptualization, Methodology, Resources, Supervision, Validation, Writing - Review \& Editing, Project administration, Funding acquisition.


\begin{thebibliography}{}

\bibitem[Aliakbarian et~al., 2017]{Aliakbarian2017}
Aliakbarian, M.~S., Saleh, F.~S., Salzmann, M., Fernando, B., Petersson, L.,
  and Andersson, L. (2017).
\newblock {Encouraging LSTMs to Anticipate Actions Very Early}.
\newblock {\em Proceedings of the IEEE International Conference on Computer
  Vision}, pages 280--289.

\bibitem[André et~al., 2020]{9096121}
André, J., Lopes, J., Palermo, M., Gonçalves, D., Matias, A., Pereira, F.,
  Afonso, J., Seabra, E., Cerqueira, J., and Santos, C. (2020).
\newblock Markerless gait analysis vision system for real-time gait monitoring.
\newblock In {\em 2020 IEEE International Conference on Autonomous Robot Systems and Competitions (ICARSC)}, pages 269--274.

\bibitem[Baptista-Rios et~al., 2020]{Baptista-Rios2020}
Baptista-Rios, M., Lopez-Sastre, R.~J., {Caba Heilbron}, F., {Van Gemert},
  J.~C., Acevedo-Rodriguez, F.~J., and Maldonado-Bascon, S. (2020).
\newblock {Rethinking Online Action Detection in Untrimmed Videos: A Novel Online Evaluation Protocol}.
\newblock {\em IEEE Access}, 8:5139--5146.

\bibitem[Berardini et~al., 2020]{Berardini2020}
Berardini, D., Moccia, S., Migliorelli, L., Pacifici, I., di~Massimo, P.,
  Paolanti, M., and Frontoni, E. (2020).
\newblock {Fall detection for elderly-people monitoring using learned features and recurrent neural networks}.
\newblock {\em Experimental Results}, 1:1--9.

\bibitem[Bonney et~al., 2016]{Bonney2016}
Bonney, H., de~Silva, R., Glunti, P., Greenfield, J., and Hunt, B. (2016).
\newblock {Management of the ataxias towards best clinical practice Third edition}.
\newblock (July):29.

\bibitem[Canuto et~al., 2021]{Canuto2021}
Canuto, C., Moreno, P., Samatelo, J., Vassallo, R., and Santos-Victor, J.
  (2021).
\newblock {Action anticipation for collaborative environments: The impact of contextual information and uncertainty-based prediction}.
\newblock {\em Neurocomputing}, 444:301--318.

\bibitem[Celik et~al., 2021]{Celik2021}
Celik, Y., Stuart, S., Woo, W.~L., and Godfrey, A. (2021).
\newblock {Gait analysis in neurological populations: Progression in the use of wearables}.
\newblock {\em Medical Engineering and Physics}, 87:9--29.

\bibitem[Chalen and Vintimilla, 2019]{Chalen2019}
Chalen, T.~M. and Vintimilla, B. (2019).
\newblock {Towards Action Prediction Applying Deep Learning}.
\newblock {\em 2019 IEEE Latin American Conference on Computational Intelligence, LA-CCI 2019}, pages 1--3.

\bibitem[Chalvatzaki et~al., 2020]{Chalvatzaki2020}
Chalvatzaki, G., Koutras, P., Tsiami, A., Tzafestas, C.~S., and Maragos, P.
  (2020).
\newblock {\em {i-Walk Intelligent Assessment System: Activity, Mobility, Intention, Communication}}, volume 12538 LNCS.
\newblock Springer International Publishing.

\bibitem[Chalvatzaki et~al., 2019]{Chalvatzaki2019}
Chalvatzaki, G., Papageorgiou, X.~S., Maragos, P., and Tzafestas, C.~S. (2019).
\newblock {Learn to Adapt to Human Walking: A Model-Based Reinforcement Learning Approach for a Robotic Assistant Rollator}.
\newblock {\em IEEE Robotics and Automation Letters}, 4(4):3774--3781.

\bibitem[Cheng and Wu, 2017]{Cheng2017}
Cheng, W.~C. and Wu, Y.~Z. (2017).
\newblock {A user's intention detection method for smart walker}.
\newblock {\em Proceedings - 2017 IEEE 8th International Conference on Awareness Science and Technology, iCAST 2017}, pages 35--39.

\bibitem[{De Geest} and Tuytelaars, 2018]{DeGeest2018}
{De Geest}, R. and Tuytelaars, T. (2018).
\newblock {Modeling temporal structure with LSTM for online action detection}.
\newblock {\em Proceedings - 2018 IEEE Winter Conference on Applications of Computer Vision, WACV 2018}, pages 1549--1557.

\bibitem[Deng et~al., 2010]{Deng2010}
Deng, J., Dong, W., Socher, R., Li, L.-J., {Kai Li}, and {Li Fei-Fei} (2010).
\newblock {ImageNet: A large-scale hierarchical image database}.
\newblock pages 248--255.

\bibitem[Figueiredo et~al., 2020]{Figueiredo2020}
Figueiredo, J., Carvalho, S.~P., Goncalve, D., Moreno, J.~C., and Santos, C.~P. (2020).
\newblock {Daily Locomotion Recognition and Prediction: A Kinematic Data-Based Machine Learning Approach}.
\newblock {\em IEEE Access}, 8:33250--33262.

\bibitem[Gao et~al., 2017]{Gao2017}
Gao, J., Yang, Z., and Nevatia, R. (2017).
\newblock {Red: Reinforced encoder-decoder networks for action anticipation}.
\newblock {\em British Machine Vision Conference 2017, BMVC 2017}.

\bibitem[Girdhar et~al., 2019]{Girdhar2019}
Girdhar, R., {Joao Carreira}, J., Doersch, C., and Zisserman, A. (2019).
\newblock {Video action transformer network}.
\newblock {\em Proceedings of the IEEE Computer Society Conference on Computer Vision and Pattern Recognition}, pages 244--253.

\bibitem[Guo et~al., 2020]{Guo2020}
Guo, S., Qing, L., Miao, J., and Duan, L. (2020).
\newblock {Action prediction via deep residual feature learning and weighted
  loss}.
\newblock {\em Multimedia Tools and Applications}, 79(7-8):4713--4727.

\bibitem[He et~al., 2016]{He2016}
He, K., Zhang, X., Ren, S., and Sun, J. (2016).
\newblock {Deep residual learning for image recognition}.
\newblock {\em Proceedings of the IEEE Computer Society Conference on Computer Vision and Pattern Recognition}, pages 770--778.

\bibitem[Huang et~al., 2005]{Huang2005}
Huang, C., Wasson, G., Alwan, M., Sheth, P., and Ledoux, A. (2005).
\newblock {Shared navigational control and user intent detection in an
  intelligent walker}.
\newblock {\em AAAI Fall Symposium - Technical Report}, pages 59--66.

\bibitem[Jalal et~al., 2017]{Jalal2017}
Jalal, A., Kim, Y.~H., Kim, Y.~J., Kamal, S., and Kim, D. (2017).
\newblock {Robust human activity recognition from depth video using
  spatiotemporal multi-fused features}.
\newblock {\em Pattern Recognition}, 61:295--308.

\bibitem[Jim{\'{e}}nez et~al., 2019]{Jimenez2019}
Jim{\'{e}}nez, M.~F., Monllor, M., Frizera, A., Bastos, T., Roberti, F., and
  Carelli, R. (2019).
\newblock {Admittance Controller with Spatial Modulation for Assisted
  Locomotion using a Smart Walker}.
\newblock {\em Journal of Intelligent and Robotic Systems: Theory and
  Applications}, 94(3-4):621--637.

\bibitem[Jonsdottir and Ferrarin, 2018]{Jonsdottir2018}
Jonsdottir, J. and Ferrarin, M. (2018).
\newblock {Gait disorders in persons after stroke}.
\newblock {\em Handbook of Human Motion}, 2-3:1205--1216.

\bibitem[Ke et~al., 2019]{Ke2019}
Ke, Q., Fritz, M., and Schiele, B. (2019).
\newblock {Time-conditioned action anticipation in one shot}.
\newblock {\em Proceedings of the IEEE Computer Society Conference on Computer Vision and Pattern Recognition}, pages 9917--9926.

\bibitem[Kozlov et~al., 2020]{Kozlov2020}
Kozlov, A., Andronov, V., and Gritsenko, Y. (2020).
\newblock {Lightweight network architecture for real-time action recognition}.
\newblock {\em Proceedings of the ACM Symposium on Applied Computing}, pages 2074--2080.

\bibitem[Kurai et~al., 2019]{Kurai2019}
Kurai, T., Shioi, Y., Makino, Y., and Shinoda, H. (2019).
\newblock {Temporal conditions suitable for predicting human motion in
  walking}.
\newblock {\em Conference Proceedings - IEEE International Conference on
  Systems, Man and Cybernetics}, pages 2986--2991.

\bibitem[Li et~al., 2016]{Li2016}
Li, Y., Lan, C., Xing, J., Zeng, W., Yuan, C., and Liu, J. (2016).
\newblock {Online human action detection using joint classification-regression recurrent neural networks}.
\newblock {\em Lecture Notes in Computer Science (including subseries Lecture Notes in Artificial Intelligence and Lecture Notes in Bioinformatics)}, pages 203--220.

\bibitem[Liu et~al., 2019]{Liu2019}
Liu, D., Wang, Y., and Kato, J. (2019).
\newblock {Supervised spatial transformer networks for attention learning in fine-grained action recognition}.
\newblock {\em VISIGRAPP 2019 - Proceedings of the 14th International Joint Conference on Computer Vision, Imaging and Computer Graphics Theory and Applications}, pages 311--318.

\bibitem[Lopes et~al., 2021]{Lopes2021}
Lopes, J.~M., Andr{\'{e}}, J., Pereira, A., Palermo, M., Ribeiro, N.,
  Cerqueira, J., and Santos, C.~P. (2021).
\newblock {ASBGo: A Smart Walker for Ataxic Gait and Posture Assessment, Monitoring, and Rehabilitation}.
\newblock {\em Robotic Technologies in Biomedical and Healthcare Engineering}, pages 51--86.

\bibitem[Lv et~al., 2020]{LvL.YangJ.ZhaoD.2020}
Lv, L., Yang, J., Zhao, D., and Wang, S. (2020).
\newblock {A Novel Non-contact Recognition Approach of Walking Intention Based on Long Short-Term Memory Network}.
\newblock {\em Qian J., Liu H., Cao J., Zhou D. (eds) Robotics and Rehabilitation Intelligence. ICRRI 2020. Communications in Computer and Information Science}, 1335.

\bibitem[Mikolajczyk et~al., 2018]{Mikolajczyk2018}
Mikolajczyk, T., Ciobanu, I., Badea, d.~I., Iliescu, A., Pizzamiglio, S., Schauer, T., Seel, T., Seiciu, P.~L., Turner, D.~L., and Berteanu, M. (2018).
\newblock {Advanced technology for gait rehabilitation: An overview}.
\newblock {\em Advances in Mechanical Engineering}, 10(7):1--19.

\bibitem[Milne et~al., 2017]{Milne2017}
Milne, S.~C., Corben, L.~A., Georgiou-Karistianis, N., Delatycki, M.~B., and Yiu, E.~M. (2017).
\newblock {Rehabilitation for Individuals with Genetic Degenerative Ataxia: A Systematic Review}.
\newblock {\em Neurorehabilitation and Neural Repair}, 31(7):609--622.

\bibitem[Moreira et~al., 2019]{Moreira2019}
Moreira, R., Alves, J., Matias, A., and Santos, C.~P. (2019).
\newblock {\em {Smart and Assistive Walker - ASBGo: Rehabilitation Robotics: A Smart-Walker to Assist Ataxic Patients}}, pages 37--68.
\newblock Springer Nature Switzerland AG.

\bibitem[M{\"{u}}ller et~al., 2017]{Muller2017}
M{\"{u}}ller, B., Ilg, W., Giese, M.~A., and Ludolph, N. (2017).
\newblock {Validation of enhanced kinect sensor based motion capturing for gait assessment}.
\newblock {\em PLoS ONE}, 12(4):14--16.

\bibitem[O'Callaghan et~al., 2020]{OCallaghan2020}
O'Callaghan, B.~P., Doheny, E.~P., Goulding, C., Fortune, E., and Lowery, M.~M. (2020).
\newblock {Adaptive gait segmentation algorithm for walking bout detection using tri-axial accelerometers}.
\newblock {\em Proceedings of the Annual International Conference of the IEEE Engineering in Medicine and Biology Society, EMBS}, pages 4592--4595.

\bibitem[Pachi and Ji, 2005]{Pachi2005FREQUENCYAV}
Pachi, A. and Ji, T. (2005).
\newblock Frequency and velocity of people walking.
\newblock {\em The Structural engineer}, 83.

\bibitem[Page et~al., 2015]{Page2015}
Page, S., Martins, M.~M., Saint-Bauzel, L., Santos, C.~P., and Pasqui, V (2015).
\newblock {Fast embedded feet pose estimation based on a depth camera for smart walker}.
\newblock {\em Proceedings - IEEE International Conference on Robotics and Automation}, pages 4224--4229.

\bibitem[Palermo et~al., 2021]{Palermo2021}
Palermo, M., Moccia, S., Migliorelli, L., Frontoni, E., and Santos, C.~P. (2021).
\newblock {Real-time human pose estimation on a smart walker using convolutional neural networks}.
\newblock {\em Expert Systems with Applications}, 184:1--15.

\bibitem[Park et~al., 2019]{Park2019}
Park, J.~H., Park, B.~O., and Lee, W.~G. (2019).
\newblock {Parametric Design and Analysis of the Arc Motion of a User-Interactive Rollator Handlebar with Hall Sensors}.
\newblock {\em International Journal of Precision Engineering and Manufacturing}, 20(11):1979--1988.

\bibitem[Paulo et~al., 2015]{Paulo2015}
Paulo, J., Peixoto, P., and Nunes, U. (2015).
\newblock {A novel vision-based human-machine interface for a robotic walker framework}.
\newblock {\em Proceedings - IEEE International Workshop on Robot and Human Interactive Communication}, pages 134--139.

\bibitem[Paulo et~al., 2017]{Paulo2017}
Paulo, J., Peixoto, P., and Nunes, U.~J. (2017).
\newblock {ISR-AIWALKER: Robotic Walker for Intuitive and Safe Mobility Assistance and Gait Analysis}.
\newblock {\em IEEE Transactions on Human-Machine Systems}, 47(6):1110--1122.

\bibitem[Qiao et~al., 2020]{Qiao2020}
Qiao, Y., Cui, W., and Shi, T. (2020).
\newblock {LaM-2SRN: A method which can enhance local features and detect moving objects for action recognition}.
\newblock {\em IEEE Access}, 8:192703--192712.

\bibitem[Rodriguez-losada, 2008]{Rodriguez-losada2008}
Rodriguez-losada, D. (2008).
\newblock {A Smart Walker for the Blind}.
\newblock {\em Robotics {\&} automation Magazine}, pages 75--83.

\bibitem[Ronneberger et~al., 2015]{Ronneberger2015}
Ronneberger, O., Fischer, P., and Brox, T. (2015).
\newblock {U-Net : Convolutional Networks for Biomedical Image Segmentation}.
\newblock In {\em Medical Image Computing and Computer-Assisted Intervention
  (MICCAI)}, volume 9351 LNCS, pages 234--241. Springer.

\bibitem[Selvaraju et~al., 2019]{Selvaraju2019}
Selvaraju, R.~R., Cogswell, M., Das, A., Vedantam, R., Parikh, D., and Batra, D. (2019).
\newblock {Grad-CAM: Visual Explanations from Deep Networks via Gradient-Based Localization}.
\newblock {\em International Journal of Computer Vision}, 128(2):336--359.

\bibitem[Shen et~al., 2020]{Shen2020}
Shen, T., Afsar, M.~R., Zhang, H., Ye, C., and Shen, X. (2020).
\newblock {A 3D Computer Vision-Guided Robotic Companion for Non-Contact Human Assistance and Rehabilitation}.
\newblock {\em Journal of Intelligent and Robotic Systems: Theory and
  Applications}, 100(3-4):911--923.

\bibitem[Shi et~al., 2018]{Shi2018}
Shi, Y., Fernando, B., and Hartley, R. (2018).
\newblock {\em {Action anticipation with RBF kernelized feature mapping RNN}}, volume 11214 LNCS.
\newblock Springer International Publishing.

\bibitem[Sierra et~al., 2018]{Sierra2018}
Sierra, S.~D., Molina, J.~F., Gomez, D.~A., Munera, M.~C., and Cifuentes, C.~A. (2018).
\newblock {Development of an Interface for Human-Robot Interaction on a Robotic Platform for Gait Assistance: AGoRA Smart Walker}.
\newblock {\em 2018 IEEE ANDESCON, ANDESCON 2018 - Conference Proceedings}.

\bibitem[Simonyan and Zisserman, 2014]{Simonyan2014}
Simonyan, K. and Zisserman, A. (2014).
\newblock {Two-stream convolutional networks for action recognition in videos}.
\newblock {\em Advances in Neural Information Processing Systems}, pages 568--576.

\bibitem[Spenko et~al., 2006]{Spenko2006}
Spenko, M., Yu, H., and Dubowsky, S. (2006).
\newblock {Robotic personal aids for mobility and monitoring for the elderly}.
\newblock {\em IEEE Transactions on Neural Systems and Rehabilitation Engineering}, 14(3):344--351.

\bibitem[Vondrick et~al., 2016]{Vondrick2016}
Vondrick, C., Pirsiavash, H., and Torralba, A. (2016).
\newblock {Anticipating visual representations from unlabeled video}.
\newblock {\em Proceedings of the IEEE Computer Society Conference on Computer Vision and Pattern Recognition}, pages 98--106.

\bibitem[Weon and Lee, 2018]{Weon2018}
Weon, I.~S. and Lee, S.~G. (2018).
\newblock {Intelligent robotic walker with actively controlled human interaction}.
\newblock {\em ETRI Journal}, 40(4):522--530.

\bibitem[{WHO}, 2011]{WorldHealthOrganization2011}
{WHO} (2011).
\newblock {World report on disability.}
\newblock {\em Disability and rehabilitation}, 33(17-18):1491.

\bibitem[{WHO}, 2021]{WorldHealthOrganization}
{WHO} (2021).
\newblock {Disability and health}, Retrieved from https://www.who.int/news-room/fact-sheets/detail/disability-and-health.

\bibitem[Wu et~al., 2020]{Wu2020}
Wu, H., Ma, X., and Li, Y. (2020).
\newblock {Convolutional Networks with Channel and STIPs Attention Model for Action Recognition in Videos}.
\newblock {\em IEEE Transactions on Multimedia}, 22(9):2293--2306.

\bibitem[Yeaser et~al., 2020]{Yeaser2020}
Yeaser, A., Tung, J., Huissoon, J., and Hashemi, E. (2020).
\newblock {Learning-Aided User Intent Estimation for Smart Rollators}.
\newblock {\em Proceedings of the Annual International Conference of the IEEE Engineering in Medicine and Biology Society, EMBS}, pages 3178--3183.

\bibitem[Zhao et~al., 2020]{Zhao2020}
Zhao, X., Zhu, Z., Liu, M., Zhao, C., Zhao, Y., Pan, J., Wang, Z., and Wu, C. (2020).
\newblock {A Smart Robotic Walker With Intelligent Close-Proximity Interaction Capabilities for Elderly Mobility Safety}.
\newblock {\em Frontiers in Neurorobotics}, pages 1--17.

\bibitem[Zhou et~al., 2016]{Zhou2016}
Zhou, B., Khosla, A., Lapedriza, A., Oliva, A., and Torralba, A. (2016).
\newblock {Learning Deep Features for Discriminative Localization}.
\newblock {\em Proceedings of the IEEE Computer Society Conference on Computer Vision and Pattern Recognition}, pages 2921--2929.
\end{thebibliography}

\appendix
\section{}\label{appendix-title-4aa54719f6c4}

\subsection{Single-frame Classification Approach}
\label{appendix_classif}

\begin{figure}[H]
    \centering
    \includegraphics[width=0.9\textwidth]{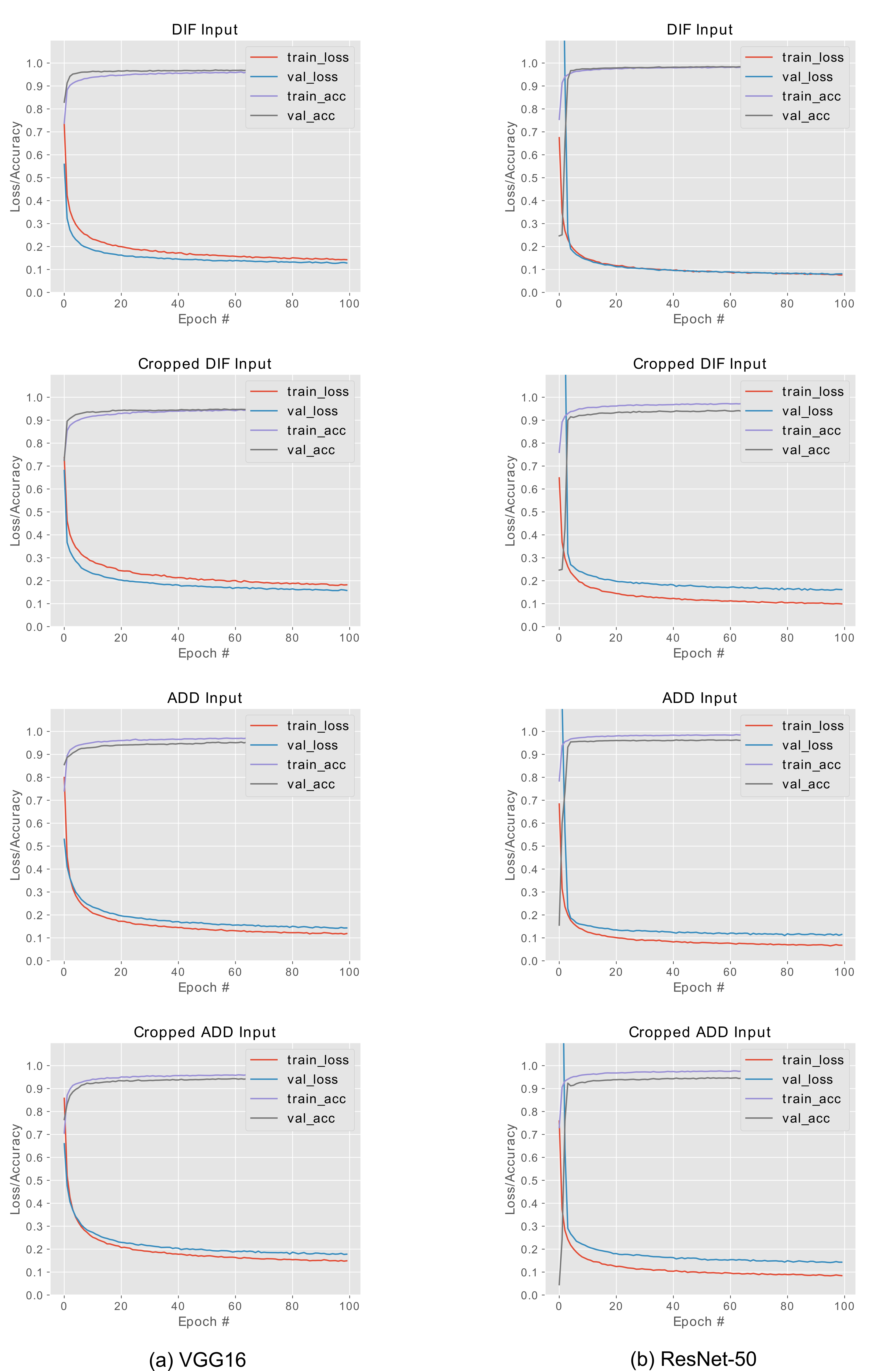}
    \caption{Accuracy and loss training curves for VGG16 and ResNet-50 models.}
    \label{fig:baselines_trainplots}
\end{figure}

Training VGG16 and ResNet-50 architectures with each developed input, computed from the acquired dataset, resulted in the training curves presented in Figure \ref{fig:baselines_trainplots}. As one can see, the overall curves are stable and with no signs of overfitting, reaching good results. It is noticeable that ResNet-50 learned faster and provided some gains in loss and accuracy.

\vspace{-1cm}
    
\begin{figure}[H]
    \centering
    \includegraphics[width=0.7\linewidth]{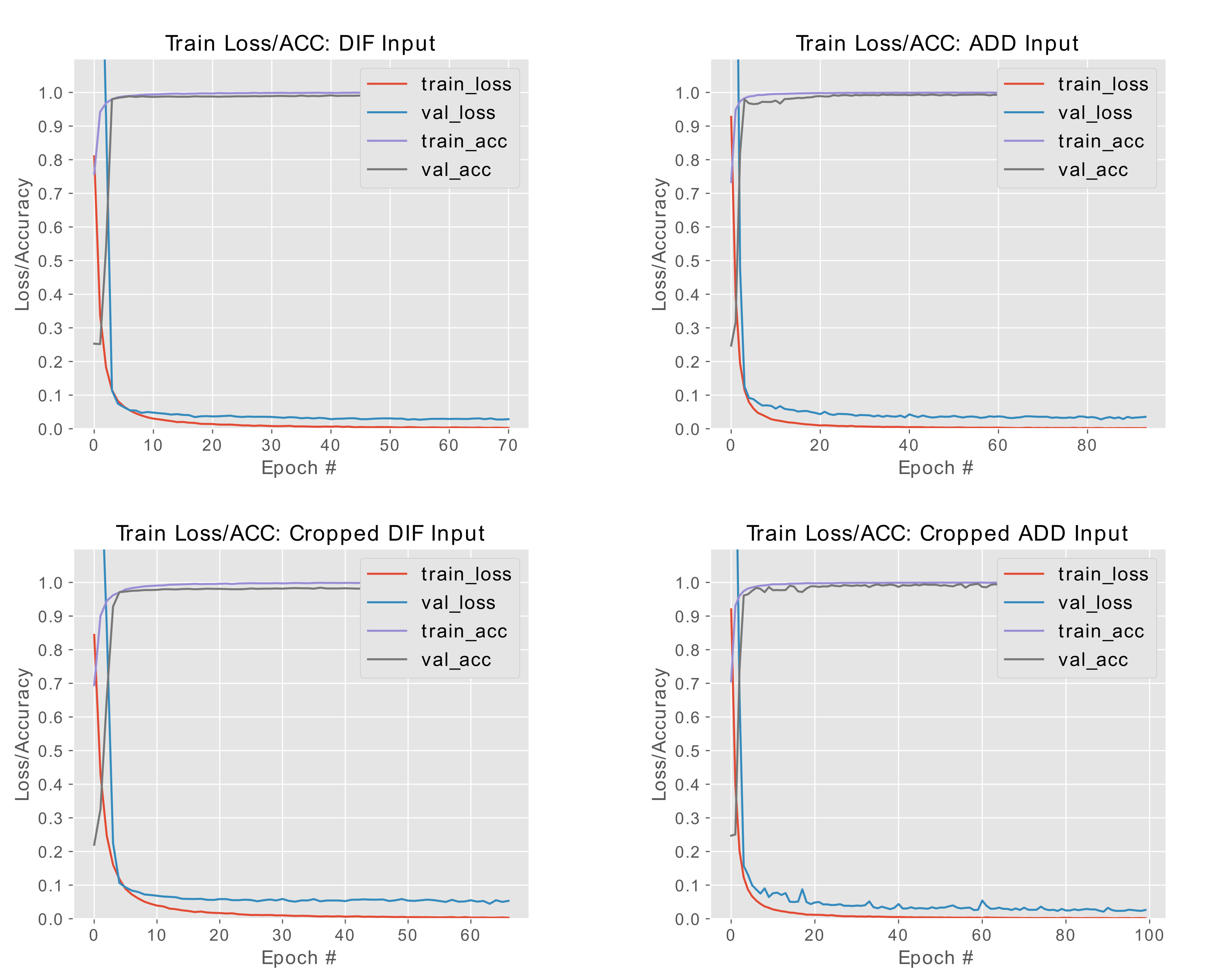}
    \caption{Accuracy and loss training curves for ResNet-50 model with an attention mechanism.}
    \label{fig:att_trainplots}
\end{figure}

Figure \ref{fig:att_trainplots} shows the obtained training curves for the ResNet-50 model with a channel-wise attention mechanism.

Table \ref{tab:test_cams} compares the grad-CAMs evaluation over the unseen test set for the VGG16 and ResNet-50 (without and with attention) models.

\begin{table}[H]
\centering
\caption{Quantative evaluation results, in percentage, for test grad-CAMs, when predicting with the VGG16 and ResNet-50 models (without and with an attention mechanism)}
\label{tab:test_cams}
\maxsizebox{0.85\textwidth}{!}{%
\begin{tabular}{@{}cc|cc|cc|cc@{}}
\toprule
\multicolumn{2}{c|}{\textbf{Input}} & \multicolumn{2}{c|}{\textbf{VGG16}} & \multicolumn{2}{c|}{\textbf{ResNet-50}} & \multicolumn{2}{c}{\textbf{ResNet-50 with attention}} \\ \midrule
\textbf{Type} & \textbf{Crop} & \textbf{\begin{tabular}[c]{@{}c@{}}Mean \\ Dice ($\pm{std}$)\end{tabular}} & \textbf{\begin{tabular}[c]{@{}c@{}}Mean \\ IoU ($\pm{std}$)\end{tabular}} & \textbf{\begin{tabular}[c]{@{}c@{}}Mean \\ Dice ($\pm{std}$)\end{tabular}} & \textbf{\begin{tabular}[c]{@{}c@{}}Mean \\ IoU ($\pm{std}$)\end{tabular}} & \textbf{\begin{tabular}[c]{@{}c@{}}Mean \\ Dice ($\pm{std}$)\end{tabular}} & \textbf{\begin{tabular}[c]{@{}c@{}}Mean \\ IoU ($\pm{std}$)\end{tabular}} \\
DIF & False & 17.18 ($\pm{8.60}$) & 9.64 ($\pm{5.31}$) & 26.01 ($\pm{15.19}$) & 15.84 ($\pm{10.24}$) & 29.97 ($\pm{14.56}$) & 18.50 ($\pm{10.26}$) \\
DIF & True & 26.39 ($\pm{13.64}$) & 15.91 ($\pm{9.12}$) & 28.38 ($\pm{13.05}$) & 17.21 ($\pm{8.92}$) & \textbf{32.38} ($\pm{10.32}$) & \textbf{19.76} ($\pm{7.20}$) \\
ADD & False & 20.99 ($\pm{7.75}$) & 11.94 ($\pm{4.91}$) & 16.93 ($\pm{14.07}$) & 9.95 ($\pm{9.19}$) & 22.14 ($\pm{13.02}$) & 13.07 ($\pm{8.59}$) \\
ADD & True & \textbf{28.62} ($\pm{10.90}$) & \textbf{17.18} ($\pm{7.64}$) & \textbf{29.37} ($\pm{12.54}$) & \textbf{17.86} ($\pm{8.87}$) & 32.30 ($\pm{8.83}$) & 19.60 ($\pm{6.37}$) \\ \bottomrule
\end{tabular}%
}
\end{table}

\subsection{Segmentation-Classification Approach}
\label{appendix_seg}

\begin{figure}[h!]
    \centering
    \includegraphics[width=0.7\linewidth]{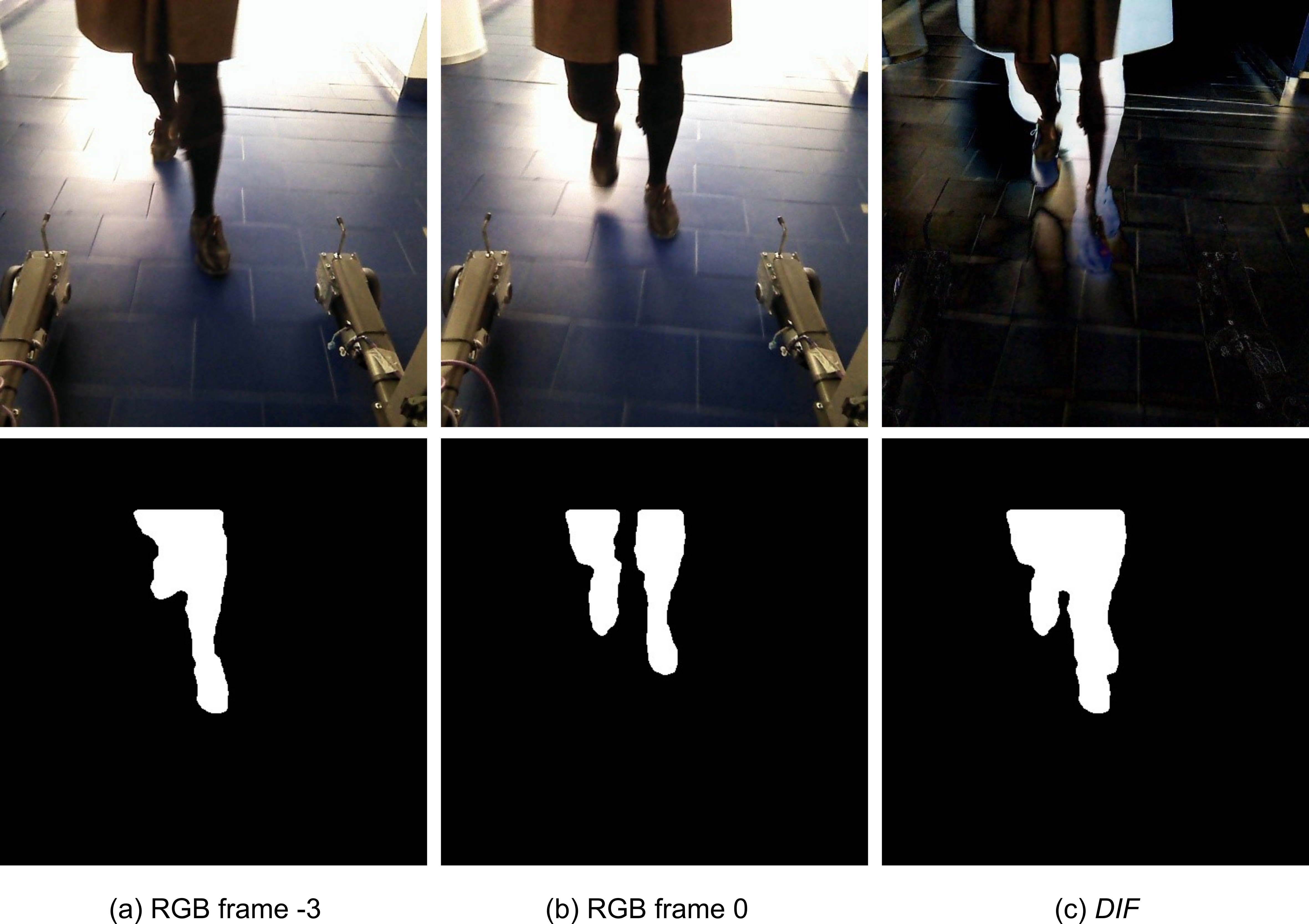}
    \caption{Examples of individually non-corrupted masks, along with their corresponding RGB inputs, for a window length of 4 frames. The presented masks are already corrected.}
    \label{fig:masks_examples}
\end{figure}

Examples of the generated masks, used as labels for segmentation and focus evaluation algorithm, are given in Figure \ref{fig:masks_examples}.

\begin{figure}[H]
    \centering
    \includegraphics[width=0.7\linewidth]{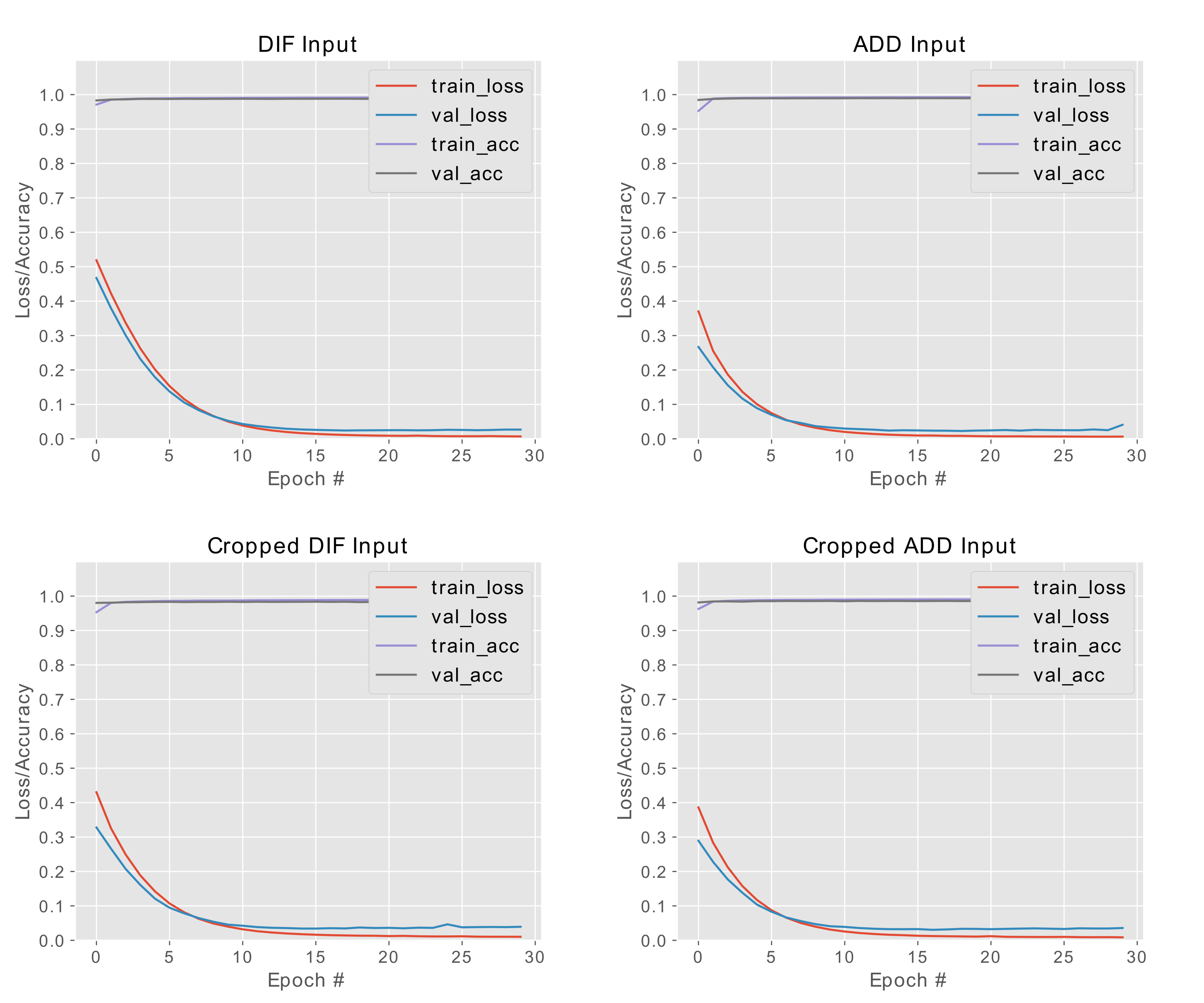}
    \caption{Accuracy and loss training curves for segmentation.}
    \label{fig:seg_trainplots}
\end{figure}

\begin{figure}[H]
    \centering
    \includegraphics[width=0.7\linewidth]{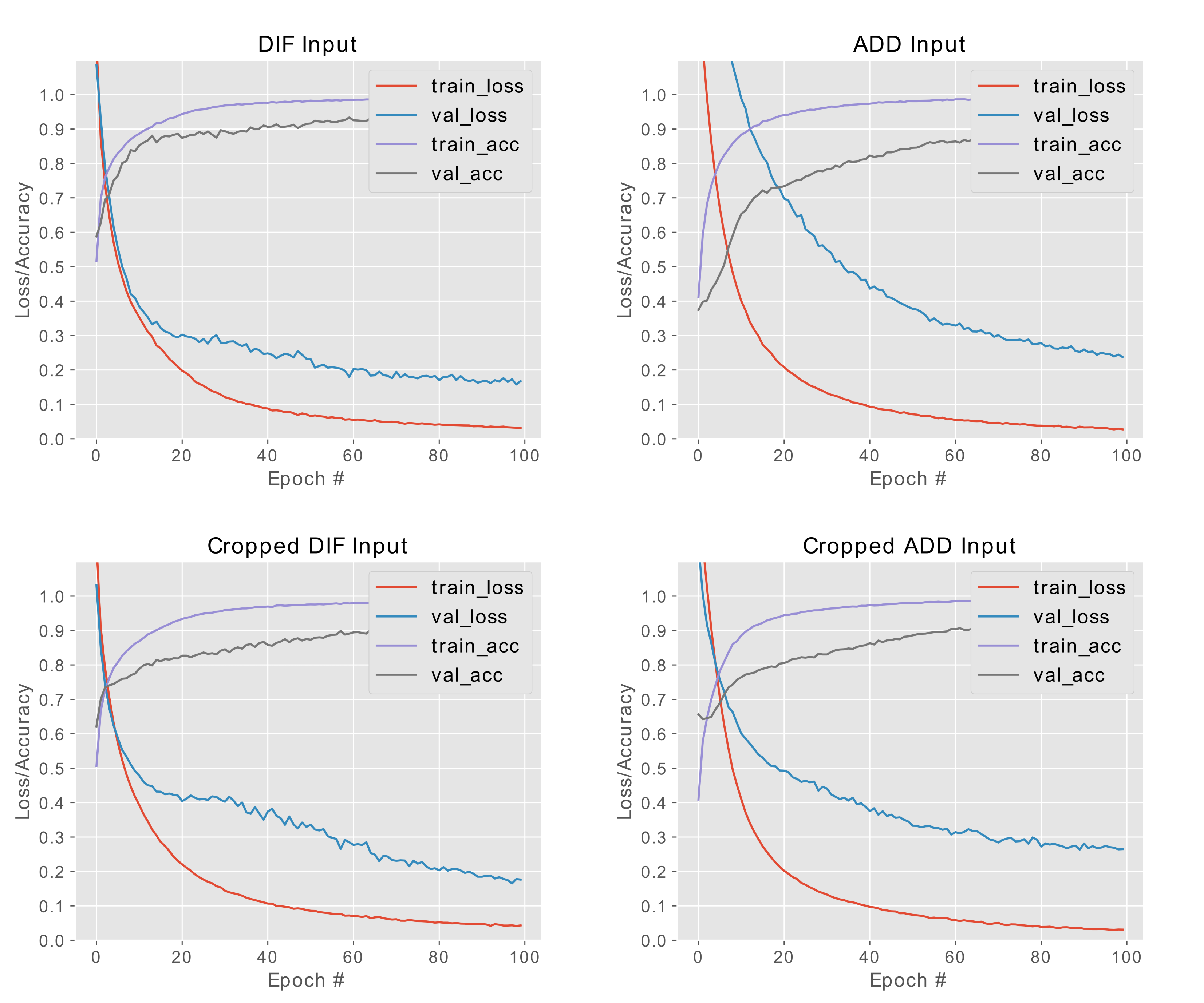}
    \caption{Accuracy and loss training curves for the adapted UNET model for classification.}
    \label{fig:segclass_trainplots}
\end{figure}

Figures \ref{fig:seg_trainplots} and \ref{fig:segclass_trainplots} display the segmentation and classification training curves, respectively, using the UNET and adapted UNET models.

\begin{figure}[H]
    \centering
    \includegraphics[width=0.7\linewidth]{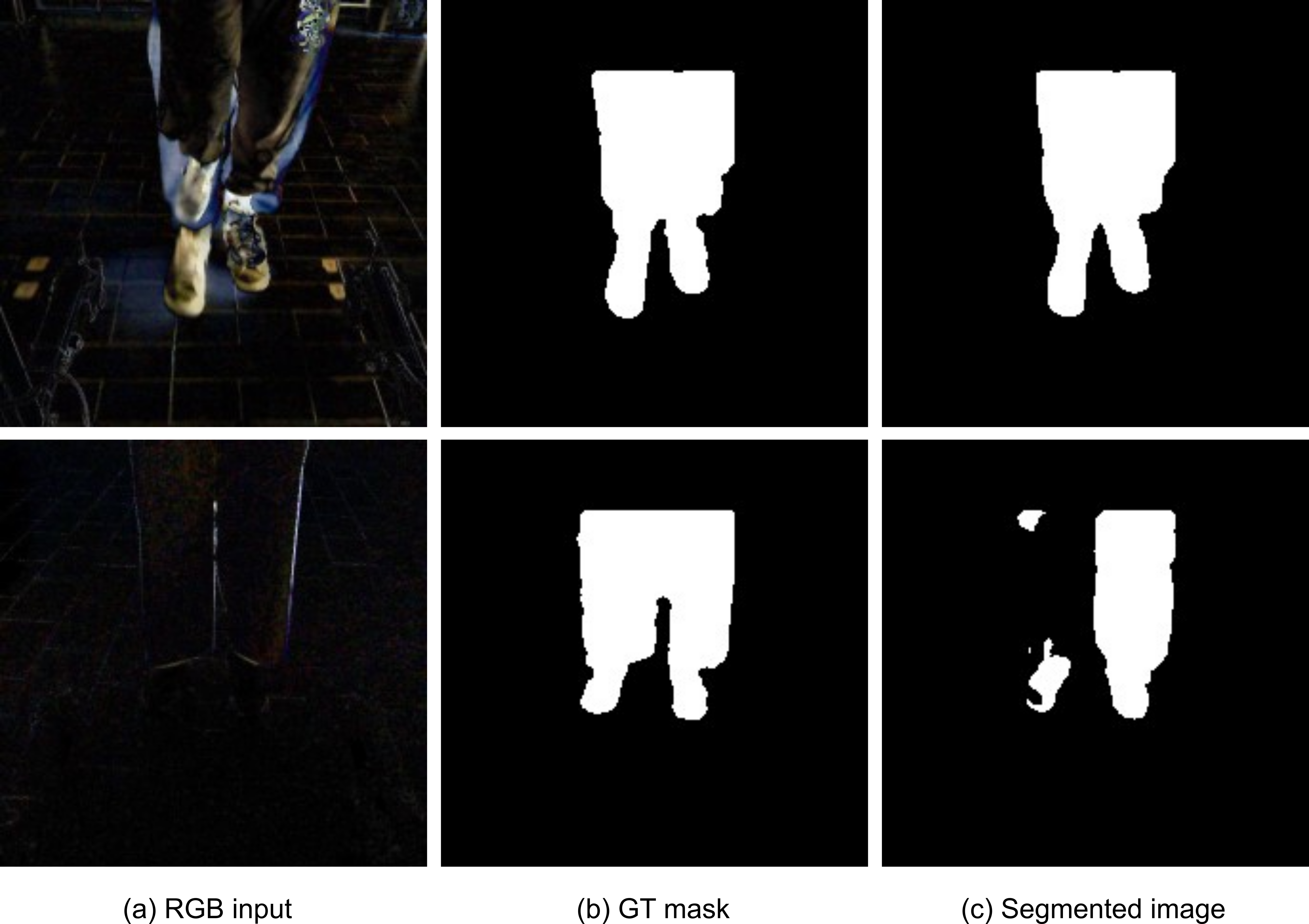}
    \caption{Examples of the best (upper) and worst (lower row) cases of segmented images, along with the respective non-cropped \emph{DIF} inputs.}
    \label{fig:seg_examples_dif}
\end{figure}

\begin{figure}[H]
    \centering
    \includegraphics[width=0.7\linewidth]{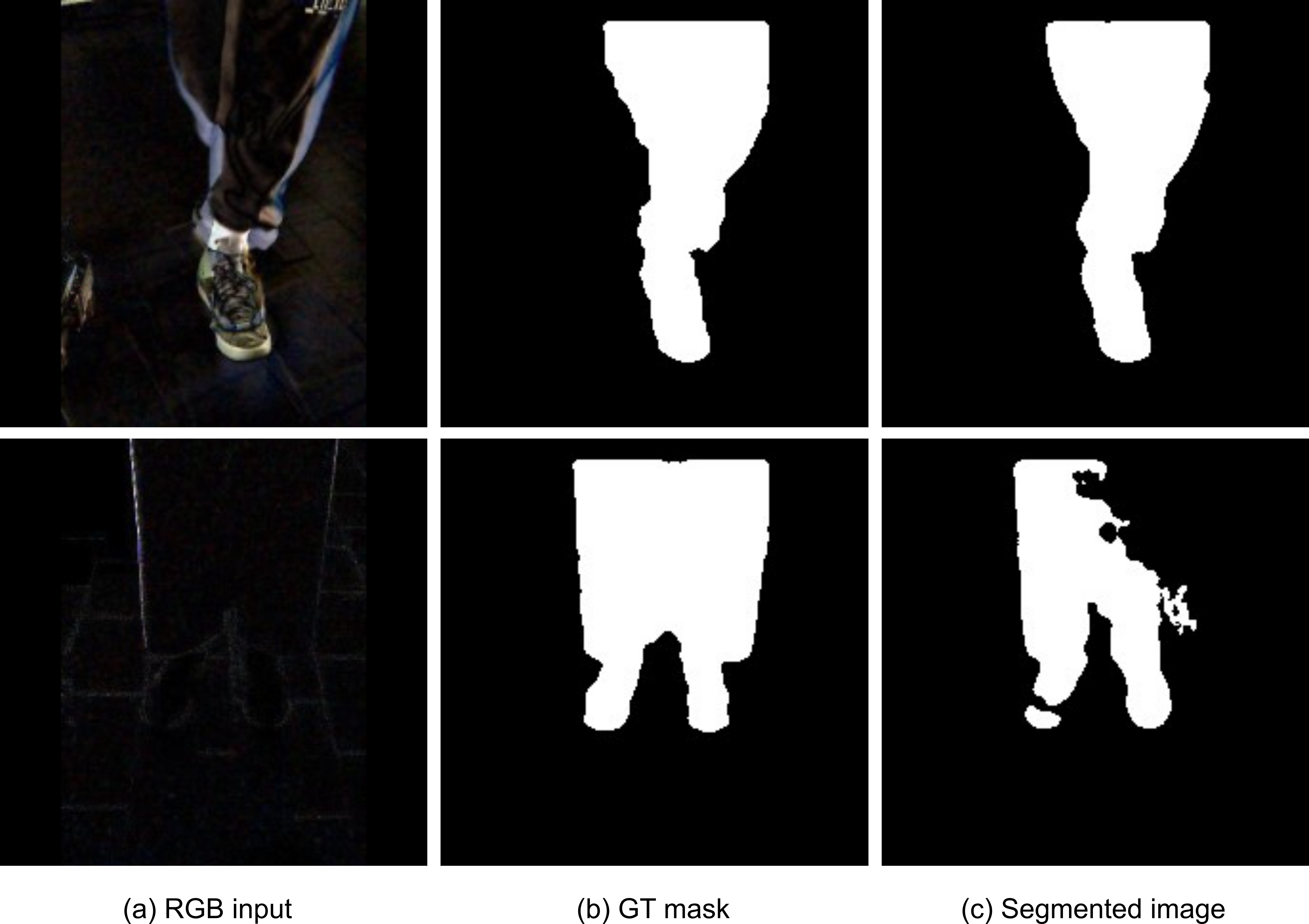}
    \caption{Examples of the best (upper) and worst (lower row) cases of segmented images, along with the respective cropped \emph{DIF} inputs.}
    \label{fig:seg_examples_difcrop}
\end{figure}

\begin{figure}[H]
    \centering
    \includegraphics[width=0.7\linewidth]{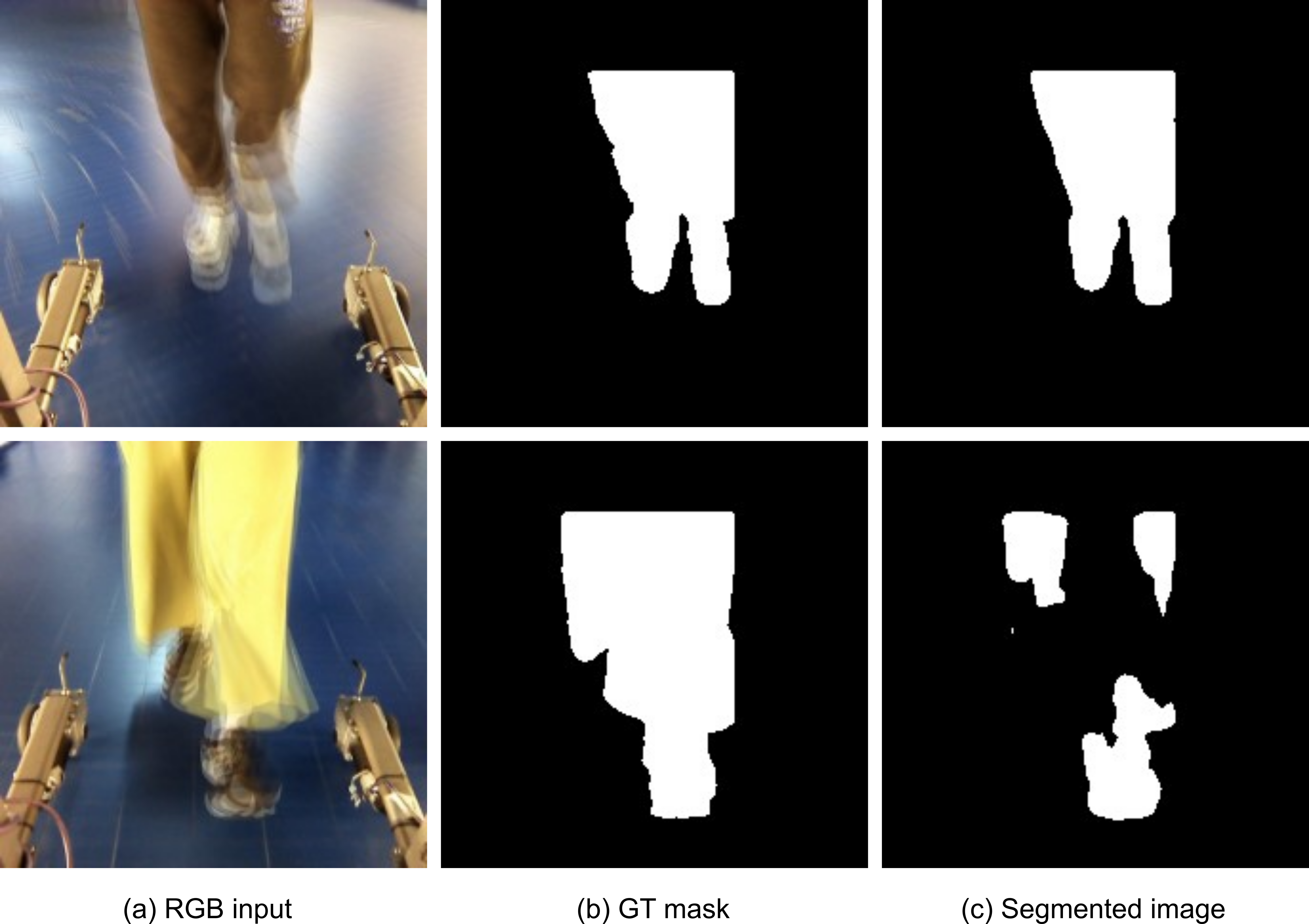}
    \caption{Examples of the best (upper) and worst (lower row) cases of segmented images, along with the respective non-cropped \emph{ADD} inputs.}
    \label{fig:seg_examples_add}
\end{figure}

\begin{figure}[H]
    \centering
    \includegraphics[width=0.7\linewidth]{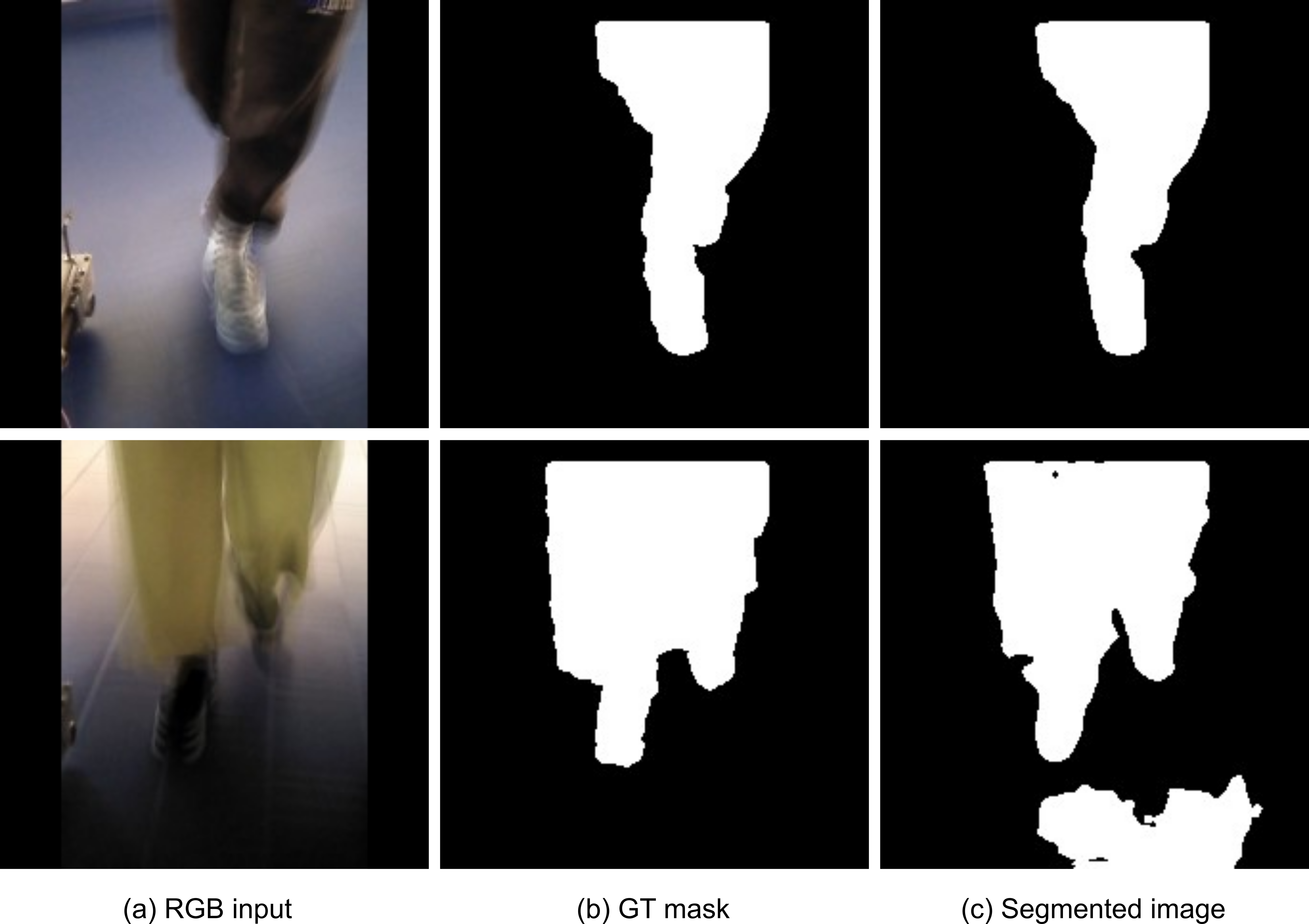}
    \caption{Examples of the best (upper) and worst (lower row) cases of segmented images, along with the respective cropped \emph{ADD} inputs.}
    \label{fig:seg_examples_addcrop}
\end{figure}

To help visualise this model's segmentation ability, Figures \ref{fig:seg_examples_dif}, \ref{fig:seg_examples_difcrop}, \ref{fig:seg_examples_add} and \ref{fig:seg_examples_addcrop} show the best and worst cases of segmented images, for each type of input. Notice that, for the cropped \emph{ADD}, the segmentation of the human body was very satisfactory, even in the worst case, although including some noise. Contrarily to this, the other inputs revealed occlusions as the apparent main factor behind a worse segmentation. The quantitative test results shown in Table \ref{tab:seg_test_quant_results} indicate that the cropped \emph{ADD} images were the easiest to segment, followed by the non-cropped \emph{ADD}, cropped \emph{DIF} and, finally, non-cropped \emph{DIF} images.

\begin{table}[H]
\centering
\caption{Evaluation results, in percentage, of the UNET model segmentation over the test set}
\label{tab:seg_test_quant_results}
\maxsizebox{0.4\textwidth}{!}{%
\begin{tabular}{@{}cccc@{}}
\toprule
\textbf{Input Type} & \textbf{Crop} & \textbf{IoU ($\pm{std}$)} & \textbf{Dice ($\pm{std}$)} \\ \midrule
DIF & False & 90.79 ($\pm{5.55}$) & 95.08 ($\pm{3.27}$) \\
DIF & True & 91.89 ($\pm{4.66}$) & 95.71 ($\pm{2.64}$) \\
ADD & False & 92.12 ($\pm{5.71}$) & 95.80 ($\pm{3.45}$) \\
ADD & True & \textbf{93.63} ($\pm{3.41}$) & \textbf{96.68} ($\pm{1.87}$)\\ \bottomrule
\end{tabular}%
}
\end{table}

\end{document}